\documentclass{article} %
\usepackage{iclr2027_conference,times}

\usepackage{amsmath,amsfonts,bm}

\def\eqref#1{equation~\ref{#1}}

\def\1{\bm{1}}

\DeclareMathAlphabet{\mathsfit}{\encodingdefault}{\sfdefault}{m}{sl}
\SetMathAlphabet{\mathsfit}{bold}{\encodingdefault}{\sfdefault}{bx}{n}

\usepackage[hyphens]{url}
\usepackage{graphicx}
\usepackage{natbib}
\setcitestyle{numbers,square,comma}
\usepackage{caption}
\usepackage{hyperref}
\usepackage{algorithm}
\usepackage{algorithmic}

\usepackage[export]{adjustbox} %
\usepackage{wrapfig}

\usepackage{arydshln}
\usepackage{amsmath} 

\usepackage{newfloat}
\usepackage{listings}
\DeclareCaptionStyle{ruled}{labelfont=normalfont,labelsep=colon,strut=off} %
\floatstyle{ruled}
\newfloat{listing}{tb}{lst}{}
\floatname{listing}{Listing}

\usepackage{booktabs}
\usepackage{multirow}

\renewcommand{\cite}{\citep}

\title{ToDoPro: Faster and Better Reference-Based Generation: Early Token Reduction via Downsampled Self-Probing}

\author{
    Caiyi Sun$^{1}$, Yujing Sun$^{2}$\thanks{Corresponding author.}, and Siu-Ming Yiu$^{1}$\\
    \textsuperscript{1}The University of Hong Kong,
    \textsuperscript{2}Nanyang Technological University\\
    \texttt{caiyi.sun@connect.hku.hk}
}
\iclrfinalcopy
\renewcommand{\lhead}[1]{}

\usepackage{xcolor}

\begin{document}

\maketitle

\begin{abstract}
Recent generic reference-based image generation models natively support multiple conditions by concatenating all reference image and text tokens into a joint sequence processed by a single Diffusion Transformer (DiT).
However, this versatile paradigm comes at an efficiency cost: accommodating multiple references leads to a token explosion, incurring massive $\mathcal{O}(N^2)$ computational complexity.
Most acceleration methods operate inside the DiT, showing limited capability in reducing spatial redundancy and failing to alleviate peak VRAM. 
While some pre-model token reduction alternatives exist, they inevitably introduce external models.
This raises a question: 
Is it possible to achieve acceleration and memory savings while preserving compatibility with in-model techniques and requiring no additional parameters?
In this work, we realize this goal by proposing the early \textbf{To}ken reduction via \textbf{Do}wnsampled self-\textbf{Pro}bing (\textbf{ToDoPro}).
By executing a few denoising steps on spatially $2\times$ downsampled reference images, this cheap probing stage utilizes the DiT model itself to effectively and efficiently extract reference token importance.
Using the probed importance, we apply a hierarchical reduction before the main denoising stage: we completely discard low-importance tokens, substitute mid-importance tokens with their downsampled counterparts to cheaply preserve global context, and keep high-importance tokens. 
Experiments across three base models and two benchmarks demonstrate that our early reduction accelerates inference significantly and mitigates peak GPU memory.
Furthermore, we can improve generation quality by safely filtering out noisy, uninformative reference tokens, enabling the model to concentrate its limited attention capacity on genuinely useful tokens.
The code will be released at \url{https://github.com/scy639/ToDoPro}
\end{abstract}

\vspace{-13pt}
\section{Introduction}
\label{sec:intro}
\vspace{-7pt}

\vspace{-3pt}
\begin{figure*}[ht]
  \begin{center}
  \includegraphics[width=0.77\columnwidth]{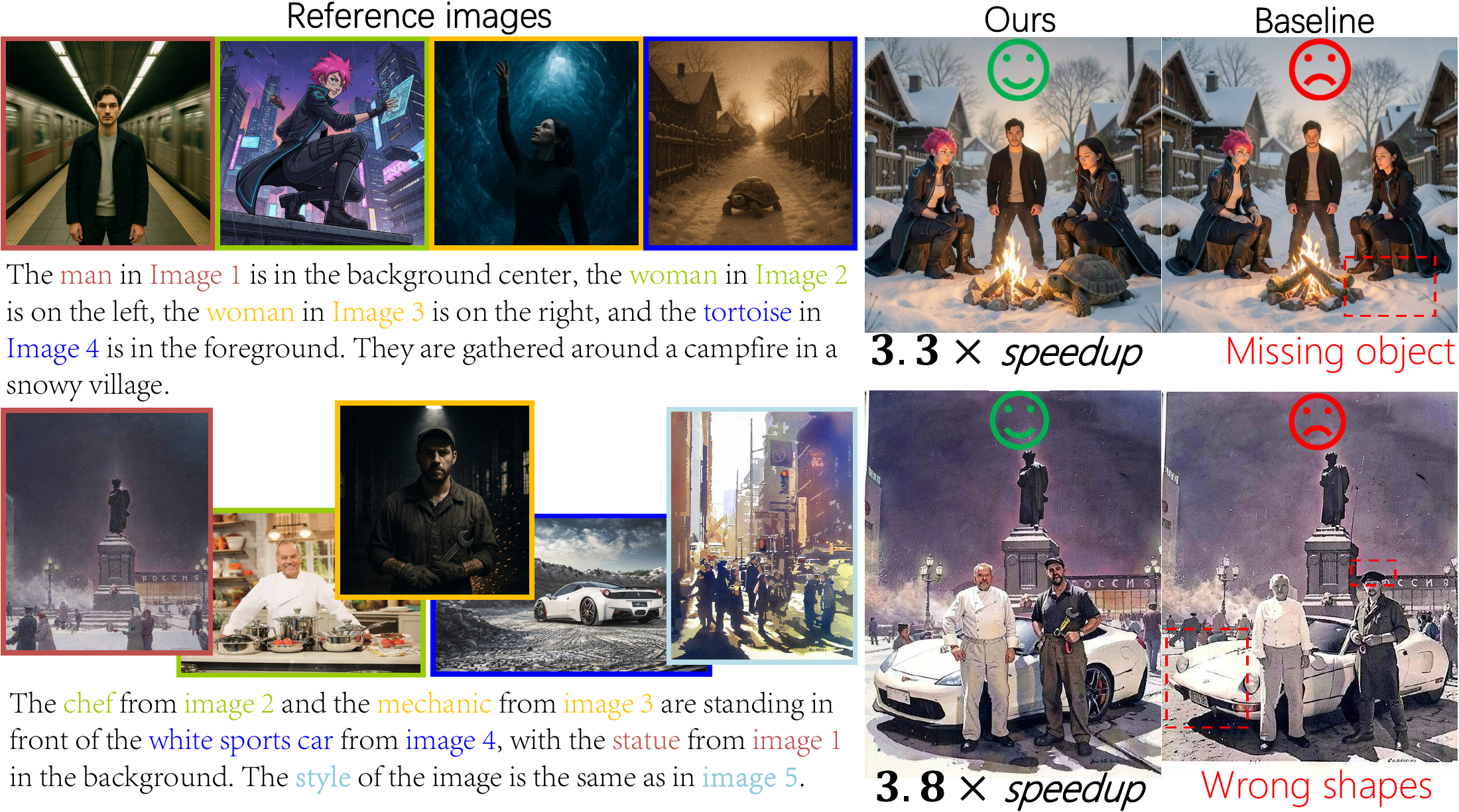}
  \end{center}
  \vspace{-7pt}
  \caption{Compared to base model FLUX.2-klein-4B, our method 
  reduces inference latency while improving generation quality.}
  \label{fig:teaser}
\end{figure*}
\vspace{-7pt}

Reference-based image generation and editing have fundamentally transformed visual synthesis, moving beyond pure text prompts to multimodal control.
By allowing multiple reference inputs, these models achieve greater controllability, driving increasing demand in domains such as advertising and design.
However, incorporating multiple reference images massively expands the token sequence length. Due to the $\mathcal{O}(N^2)$ nature of attention, this token explosion significantly increases both the \textbf{latency and peak VRAM}, limiting the number of references and resolutions that can be practically supported.
The token explosion also degrades the generation \textbf{quality} due to:
1) Attention interference: some irrelevant reference tokens inevitably attract anomalous attention, interfering with the generation process;
2) Attention dilution: The explosion in the number of tokens leads to a massive denominator in the attention Softmax, diluting weights of truly important tokens.

\begin{wrapfigure}[25]{r}{0.53\columnwidth}   %
  \centering
  \vspace{-13pt}
  \includegraphics[width=0.40\columnwidth]{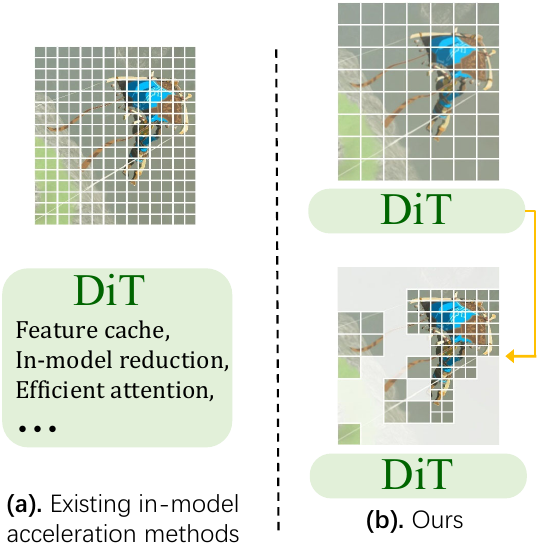}
  \vspace{-7pt}
  \caption{
(a) Most existing acceleration methods operate \textbf{inside} the DiT during the main denoising stage.
(b) ToDoPro reduces reference tokens \textbf{before} the main denoising stage begins. A probing stage on downsampled references cheaply extracts token importance, guiding token preservation / low-resolution substitution / discarding. The compressed reference images then enter the main denoising stage, slashing \textbf{end-to-end latency} and \textbf{activation VRAM} while remaining compatible with in-model methods in (a).
}
\label{fig:method_concise}
\end{wrapfigure}

Despite growing works on efficient AI, a satisfying solution for generic reference-based generation is still missing.
Most token reduction methods operate \emph{inside} the DiT, pruning or merging tokens only after the full sequence has already entered the model. Therefore, they fail to reduce peak VRAM, which typically occurs at the earliest attention layers before reduction takes place.
Some methods can reduce tokens before the main model, but they either degrade performance noticeably~\cite{tomesd}
, limiting their practicality.
Thus, to reduce peak memory and maintain compatibility with most DiT acceleration methods, we target an early reduction approach, meaning the token reduction is performed before the main denoising such that the entire main denoising process uses the reduced token sequence.

One straightforward early reduction method is to use an external model to identify unimportant regions~\cite{glimpseprune}, which introduces extra parameters and lacks alignment with the generator's actual demands.
Another method is to utilize internal features, such as attention maps, to estimate token importance. 
Yet, attention patterns for token importance remain unexplored in reference-based image generation, and insights from adjacent fields like ViTs/VLMs and text-to-image (T2I) are inapplicable to our task due to fundamental differences: 
Unlike VLMs, where visual tokens mainly provide high-level semantics, our task relies heavily on fine-grained, low-level reference details;
T2I acceleration focuses on cross-timestep temporal redundancy rather than spatial token redundancy.

Thus, we firstly perform an in-depth analysis of attention, yielding two key observations: 
First, the total attention that generative and prompt tokens pay to a reference token, termed the \emph{reference utilization score}, correlates well with its actual importance, thus serving as an effective proxy.
Second, although some unimportant reference tokens inevitably exhibit anomalous attention scores, these abnormal tokens are stochastic and vary across layers. 
Furthermore, a single layer is usually biased because different layers may have different functions and focuses.
Computing the mean utilization score across layers smooths out the bias and instantaneous noise observed in any individual layer, ensuring meaningless tokens are accurately identified, and yielding a robust importance estimation.
This reveals a pathway to address degradation: reducing the number of uninformative tokens mitigates attention dilution and interference, 
and averaging across layers further mitigates interference by identifying unimportant tokens more accurately.

However, extracting attention maps during main denoising loop cannot prevent the VRAM spike and impairs compatibility with in-model acceleration methods, contradicting the goal of early reduction.
Thus, we propose our \textbf{early token reduction via spatial-temporal downsampled self-probing}, which identifies unimportant tokens in a cheap probing denoising stage and prunes or replaces them prior to the main denoising.
By executing a temporally downsampled, few-step denoising on spatially $2\times$ downsampled images, this probing stage cheaply approximates attention of full-resolution loop.
Then, we adopt different policies for each $2 \times 2$ block of full-resolution reference tokens, guided by importance score of its corresponding downsampled reference token:
High-importance blocks remain intact (preserving all 4 tokens), whereas low-importance blocks are completely discarded. 
To prevent quality drops from losing broader mid-importance context at high pruning rates, 
we introduce an intermediate level, replacing mid-importance blocks with the corresponding downsampled reference token instead of completely discarding them, cheaply capturing global context.

This early reduction yields three major benefits: 
(1) Because tokens are reduced before the entire main denoising stage, computational savings apply to every timestep and layer, slashing FLOPs significantly and mitigating peak VRAM. 
(2) It is inherently orthogonal, complementary with existing in-model acceleration techniques (e.g., feature caching and in-model token reduction).
(3) The self-probing design utilizes the generator itself, naturally ensuring optimal alignment with the model's actual demands while avoiding external models.

More importantly, our method does not sacrifice performance for efficiency; rather, it comprehensively enhances generation quality. 
By discarding low-importance tokens, our method mitigates attention interference and dilution issues, allowing the model to better concentrate its limited attention capacity on prompt and generative tokens and genuinely informative reference tokens, ultimately improving overall generation quality.

Our key contributions are threefold:
(1)~We present the first in-depth analysis of attention in generic reference-based diffusion models. This analysis uncovers two key observations regarding reference utilization, and layer-wise bias and noise, establishing a reliable proxy for reference token importance.
(2)~Guided by these insights, we propose a training-free, early token reduction framework, which is inherently compatible with the existing ecosystem of in-model acceleration.
(3)~Experiments across three base models and two benchmarks show that our method delivers up to $3\times$ mean speedups, cuts peak activation VRAM by up to $50\%$, and consistently improves generation quality.

\section{Related Work}

\subsection{Reference-based Image Generation and Editing}

Diffusion models have achieved success in text-to-image generation, motivating extensive work extending them to reference-conditioned generation or editing.
These methods incorporate diverse visual conditions (e.g., masks, layouts, subject images, exemplar pairs, and pose maps) to achieve fine-grained controllability.
Many methods~\cite{lin2025ctrl,anycontrol,pixelponder,unicombine,Durian,unibiotransfer,gong2026relationadapter} integrated references via dedicated control encoders, adapters, or condition-selection mechanisms. 
While effective in their target tasks, they rely heavily on task-tailored architectures or input assumptions, limiting scalability.

More recently, general-purpose models have emerged to unify diverse conditioning, generation, and editing tasks within a shared DiT backbone~\cite{flux2,qwenImage,firered,A2Edit}
by concatenating reference, text, and generative tokens into a single sequence. This paradigm supports arbitrary reference counts and resolutions, covering diverse downstream tasks.
However, this generality comes at a steep cost: token counts and quadratic self-attention complexity explode with additional references.
This work focuses on this emerging general-purpose paradigm, which is more versatile and representative of future directions.

\subsection{Token Reduction for Vision Models}

Token reduction, also named as token compression, has been explored to improve vision model efficiency (including ViTs and MLLMs) by shortening the token sequence~\cite{survey_mllm_compress_2026,survey_mllm_token_compress_2026,survey_token_reduction_position,survey_lvlm_inference}.
One major category is \emph{in-model} token reduction, which progressively discards or merges tokens during the forward pass of the main model~\cite{transPrune,repreShift}.
Another category is \emph{pre-model} token reduction, which compresses tokens before entering the main model.
Among these, training-free methods like similarity-based token merging~\cite{tomesd} can achieve this goal without introducing extra trainable modules, though their performance is bounded.
Some works~\cite{visionzip,SCOPE,prumerge,hired} exploit CLS-to-token attention from a vision encoder to prune tokens before feeding them to VLMs. 
Other works~\cite{glimpseprune,flexselect} adopt more precise methods to predict region importance, but at the cost of introducing new external models.

\subsection{Efficient Image and Video Generation}

Diffusion acceleration has been investigated from multiple complementary perspectives.
Cross-step caching methods~\cite{deepcache,delta_dit,toca,teacache,tgate,l2c,pab,corgi,soda,ditfastattn,tap} reuse features or attention maps across denoising steps to eliminate redundant computation.
Distillation methods~\cite{lcm,add} reduce sampling steps, model pruning~\cite{LD_Pruner_cvpr24, structural_pruning_nips23_maXinYin, Dip_go_nips23} removes structural weights, quantization~\cite{qdiff,qdit} lowers arithmetic precision, and in-model token reduction methods~\cite{sparsedit,sdtm,diffsparse,catPrune,CoReDiT,DaTo} 
compress tokens inside models.

Another direction is efficient attention~\cite{zhou2026trainable_nlog,linfusion,sana,clear}, which develop linear or log-linear attention variants.
However, these methods only reduce computation inside attention, thus cannot save computation in Feed-Forward Networks.

\subsection{Efficient Reference-Based Generation and Editing}

Closer to our reference-based generation setting, several works accelerate specific conditional generation scenarios.
RegionE and SpotEdit~\cite{spotedit,regionE} reduce computation on unchanged regions in image editing, but they are limited to layout-preserving single-image editing setting.
OmniControl2~\cite{ominicontrol2_arxiv25} removes near-zero tokens in Canny edge maps 
and requires re-training.
EVCtrl~\cite{evctrl} reduces computation within ControlNet, and
ToPi~\cite{topi} accelerates in-context generation by pruning tokens inside the main denoising stage.

To the best of our knowledge, an early-stage reference-based generation acceleration method that reduces reference tokens before the main denoising run is still absent.
By shortening the input sequence prior to the DiT in main denoising loop, our approach is inherently orthogonal and complementary to in-model techniques (e.g., caching, quantization, efficient attention, KV sparsification, and in-model token reduction).

\section{Method}
\label{sec:method}

\subsection{Preliminaries and Notations}
\label{sec:prelim}

Generic reference-based diffusion models (e.g., Flux.2) unify visual conditions by concatenating target generative tokens $G$, text prompt tokens $P$ (from a text encoder),
and clean reference tokens $R$ (from a frozen VAE) into a joint sequence processed by an $L$-layer Diffusion Transformer (DiT). 
Let $G^{t,l}$, $P^{t,l}$, and $R^{t,l}$ denote token states at timestep $t$ and layer $l$. Denoising starts from pure noise $G^{T, 0}$ down to $G^{0, L}$ for final decoding. At step $t$, reference tokens are first modulated as $R^{t, 0} = \text{Modulate}(R, t)$. After passing through all $L$ layers, the generative output $G^{t, L}$ is updated via a numerical solver 
as $G^{t-1, 0} = \mathcal{F}_{\text{solver}} \big( G^{t, 0}, G^{t, L} \big)$.

For the importance score $I$, omitting a superscript denotes averaging over the corresponding axis: omitting $l$ denotes averaging over layers ($\frac{1}{L}\sum_l [\cdot]$), omitting $t$ denotes averaging over timesteps ($\frac{1}{T}\sum_t [\cdot]$), and omitting both denotes averaging over both axes ($\frac{1}{LT}\sum_t \sum_l [\cdot]$). For all other variables (e.g., $A$, $G$, $P$, $R$), omitting $t$ or $l$ is strictly notational brevity and does not imply averaging.

\subsection{Investigation of Attention in Reference-Based Generation}

Attention scores are widely adopted as a proxy for token importance in ViTs and VLMs, where prior works observe the rule that visual information rapidly aggregates into a few "anchor" tokens during early layers, allowing most computationally silent vision tokens to be safely pruned~\cite{visionzip,fastv}.
While visual tokens in VLMs primarily serve high-level semantic understanding, our task relies heavily on them to continuously provide fine-grained, low-level spatial details for generation. 
Thus, the rules observed in VLMs cannot be trivially transferred to our task.
Furthermore, acceleration methods in text-to-image (T2I) generation primarily focus on cross-timestep temporal feature similarity.
Given that the intermediate feature dynamics within generic reference-based generation models remain largely unexplored, we first perform an in-depth analysis of the attention patterns specific to this task.

\subsubsection{Reference Utilization Score}

\begin{wrapfigure}[16]{r}{0.49\textwidth}
  \centering
  \includegraphics[width=0.49\columnwidth]{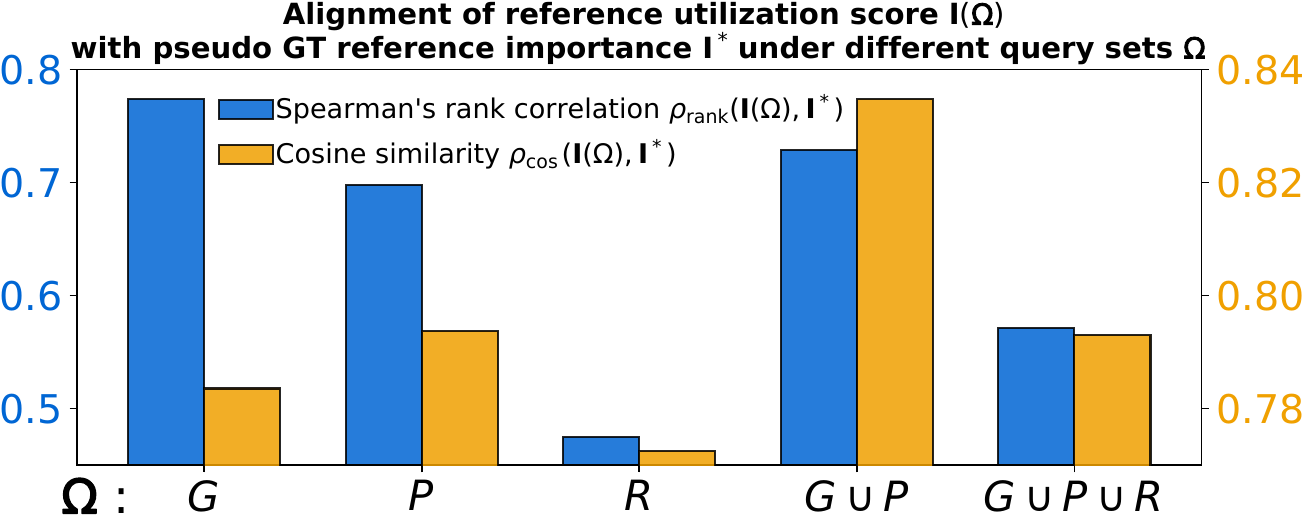}
  \caption{
Alignment of reference utilization scores with pseudo ground-truth importance across different query sets ($\Omega$).
The set ($\Omega = G \cup P$) consistently yields the best overall alignment, demonstrating that querying reference tokens via both generative and prompt tokens optimally proxies the importance.
}\label{fig:imp_query}
\end{wrapfigure}
We first explore how to quantify reference token importance using the model's intrinsic attention behaviour.
Intuitively, a reference token's importance should be proportional to the attention it attracts during the denoising pass.
Formally, let $A_{i, j}^{t, l} = \frac{1}{H} \sum_{h=1}^H q_{i, h}^{t, l} (k_{j, h}^{t, l})^\top$ denote the head-averaged attention score from query $i$ to key $j$
(for notational simplicity, we omit the normalization like Softmax here
).
For a given reference token $r$, we define the reference utilization score queried by a specific query set $\Omega$ as $I_r^{t, l}(\Omega) = \sum_{i \in \Omega} A_{i, r}^{t, l}$ and evaluate query combinations, including $\Omega \in \{G, P, G \cup P, G \cup P \cup R\}$: $\Omega=G$ (generative tokens only), $\Omega=P$ (prompt tokens only), $\Omega=G \cup P$, and $\Omega=G \cup P \cup R$.
To establish a pseudo ground-truth, we define the oracle importance $I_r^*$ as the gradient magnitude of the final decoded image pixels with respect to each reference token: 
$I_r^* = \left\| \nabla_{R_r} \bigl( \| \operatorname{Decode}(G^{0,L}) \|_2 \bigr) \right\|_2$, 
which reflects each reference token's impact on the generated result~\cite{att_flow_acl20}.
Then, we evaluate the alignment between this gold standard and our proposed attention-based proxy via Spearman's rank correlation $\rho_{\text{rank}}(\mathbf{I}(\Omega), \mathbf{I}^*)$, which evaluates relative token prioritization, and cosine similarity $\rho_{\cos}(\mathbf{I}(\Omega), \mathbf{I}^*)$.

We conduct this analysis on about 120 samples and take the average across all layers. 
As shown in Fig.~\ref{fig:imp_query}, the combined queries ($\Omega=G \cup P$) consistently yield the highest alignment, meaning the value of a reference token can be effectively approximated by how intensely it is queried by the generative target and the prompt semantics.
Intuitively, while $R \to R$ attention only captures task-agnostic reference structure, attention originating from generative tokens $G$ and prompt $P$ directly reflects actual conditioning demands for target generation.
Consequently, we adopt the reference utilization score queried by $G$ and $P$, formulated as $I_r = \frac{1}{LT}\sum_t \sum_l \sum_{i \in G \cup P} A_{i, r}^{t, l}$ (hereafter simply referred to as the Reference Utilization Score), as the proxy for the importance of a reference token $r$.

\subsubsection{Layer-Wise Analysis of Utilization Score}

\begin{figure*}[htbp]
\centering
\begin{tabular}{@{}c@{\hspace{-1pt}}c@{\hspace{1pt}}c@{\hspace{1pt}}c@{}}
(a) & \includegraphics[width=0.57\textwidth,valign=c]{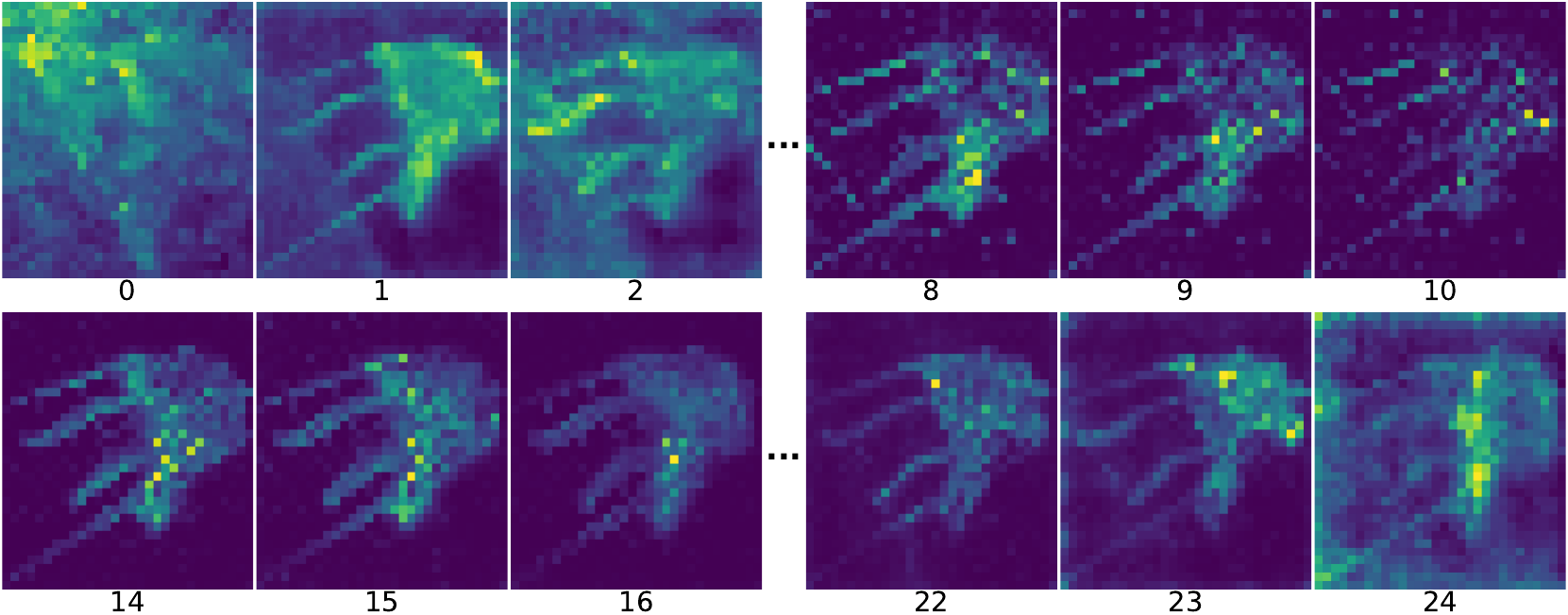}
    & (b) & \includegraphics[width=0.37\textwidth,valign=c]{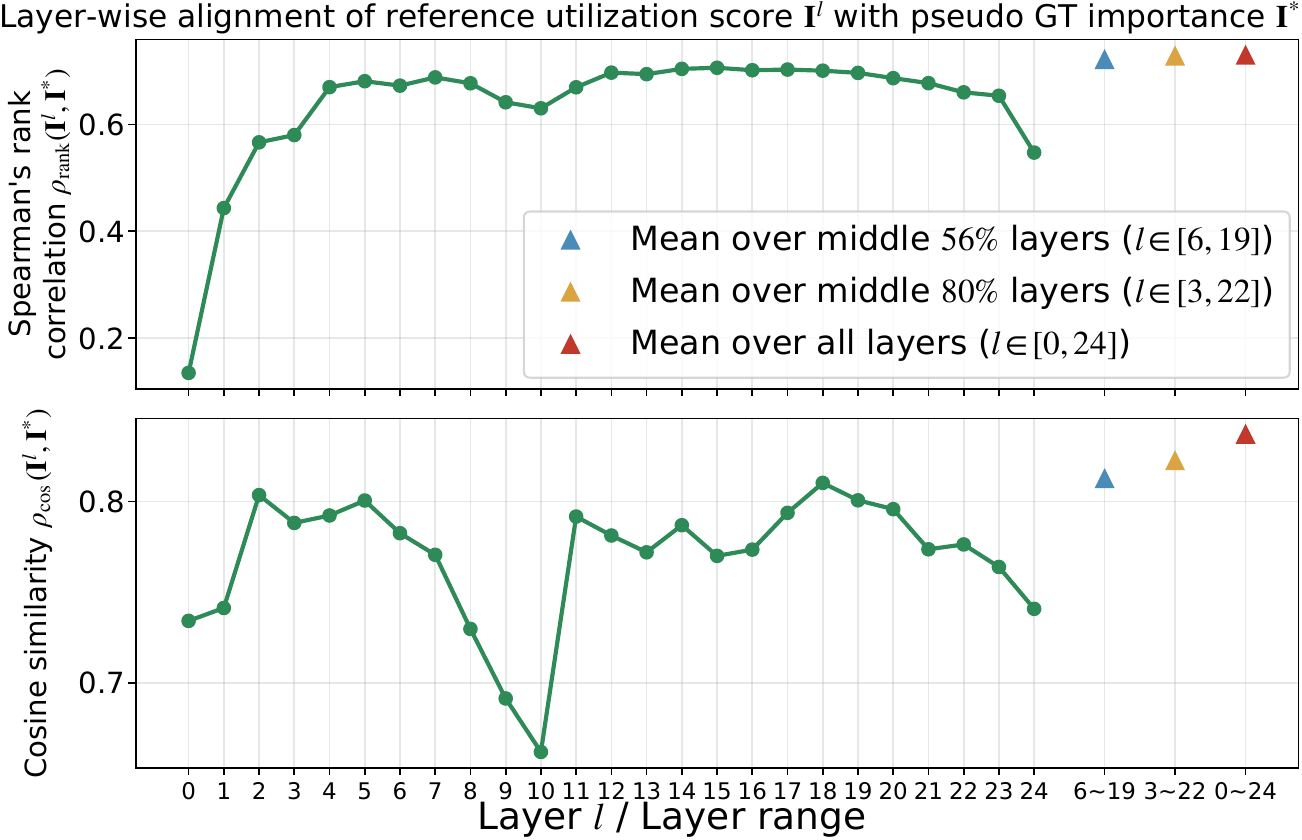}
\end{tabular}
\caption{
Analysis of layer-wise attention dynamics. (a) Layer-wise visualizations of the reference utilization score show stochastic anomalous values, and a single layer is usually biased because different layers may have different functions and focuses. (b) Computing the mean utilization score across layers smooths out the bias and instantaneous noise, achieving better alignment with the oracle token importance.
}
\label{fig:attn_vis}
\end{figure*}

As the utilization score varies across layers, we analyze its layer-wise behavior to guide the design of the final token importance estimation.
Visualizing the layer-wise attention maps of the reference utilization score $I_r^l = \frac{1}{T}\sum_t \sum_{i \in G \cup P} A_{i, r}^{t, l}$ reveals a three-stage dynamic (Fig.~\ref{fig:attn_vis}).
The early layers exhibit extensive attention, likely acting as low-level feature extractors. 
The middle layers then transition into highly concentrated patterns that strongly correlate with the actual salient regions in the references requested by the text prompt. 
The final layers show dispersed attention, decoupling from the input conditions entirely, and shift their capacity toward refining the generative tokens $G$, largely ignoring the input references.

We also show quantitative statistics in Fig.~\ref{fig:attn_vis}(b), which plots the alignment between the layer-wise reference utilization score and the oracle importance from layer 0 to 24, measured by both rank correlation (top) and cosine similarity (bottom).
Consistent with our visual observations, the middle layers exhibit better correlation compared to the early and late layers.
However, relying on any single layer remains highly susceptible to localized and anomalous token spikes.

Fortunately, as we can see in Fig.~\ref{fig:attn_vis}(a), these abnormal attention spikes are stochastic and vary across different layers.
Thus, they can effectively cancel each other out when aggregated.
As shown on the right side of Fig.~\ref{fig:attn_vis}(b), averaging the scores across multiple layers consistently outperforms any single layer.
While averaging the middle 56\% or 80\% of layers and averaging all 100\% of layers both yield excellent alignment, we simply average across all layers for better generalization across different base models.
By averaging the utilization score across all layers $I_r = \frac{1}{L}\sum_l I_r^l = \frac{1}{LT}\sum_l \sum_t \sum_{i \in G \cup P} A_{i, r}^{t, l}$, we apply an implicit low-pass filter that smooths these fluctuations and combines the different focuses of individual layers, yielding a relatively unbiased, comprehensive, and robust importance estimation.

\begin{figure*}[ht]
  \begin{center}
    \includegraphics[width=  1.0  \textwidth]{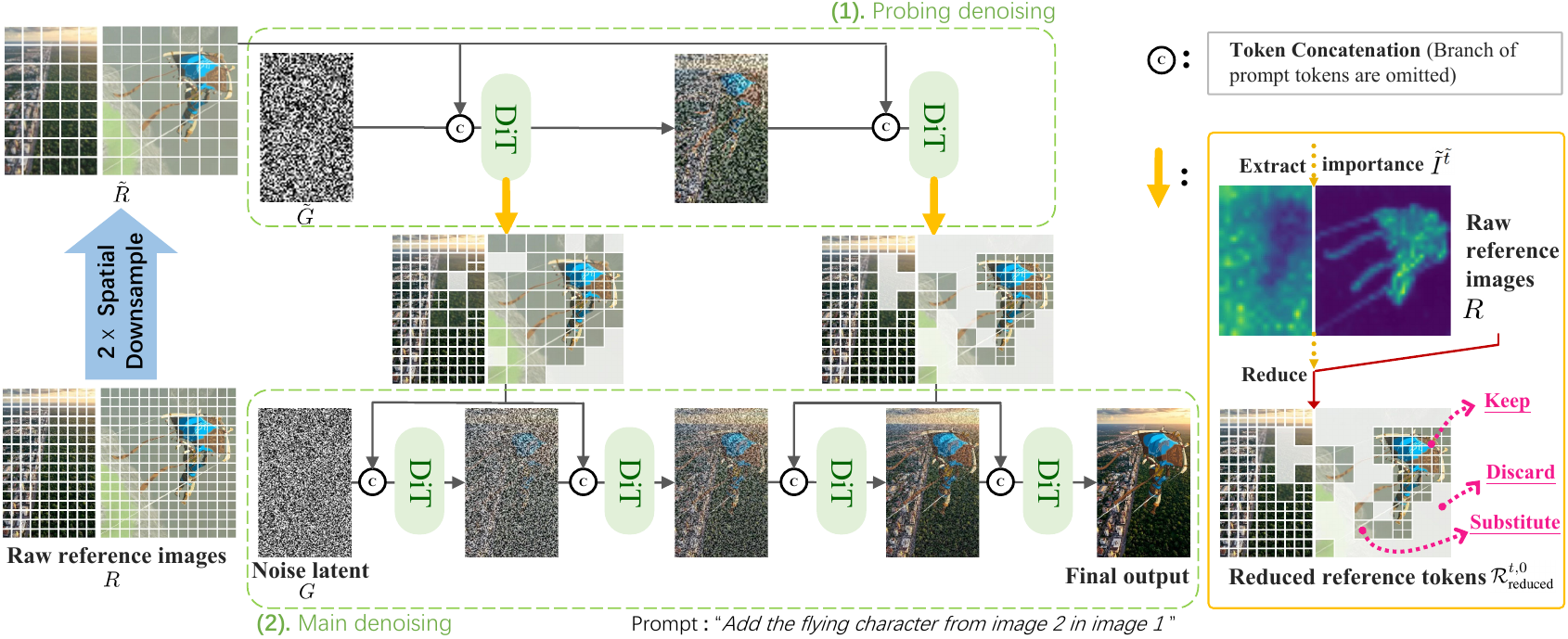}
  \end{center}
  \caption{
The illustration of our token reduction by a spatial-temporal downsampled self-probing.
By executing a downsampled time step schedule denoising loop on spatially $2\times$ downsampled input images, this probing stage cheaply extracts the reference importance. The extracted importance scores then guide the reduction strategy outside the main denoising run. The visual example is generated using FLUX.2-Klein-4B (4-step).
}
\label{fig:method}
\end{figure*}

\subsection{Early Token Reduction via Downsampled Self-Probing}

Although the full-resolution attention score is an accurate proxy for token importance, extracting it during the main denoising loop violates our early reduction goal.
Thus, we propose using a separate lightweight loop, termed the \emph{probing stage}, prior to the formal denoising stage. Unimportant reference tokens are identified and reduced before they enter the DiT in the main denoising loop.

\subsubsection{Extracting Importance via Spatial-Temporal Downsampled Self-Probing}

To cheaply identify redundant tokens, the probing stage evaluates a temporally sparse DiT denoising loop on spatially $2\times$ downsampled inputs. 
Let $\tilde{\cdot}$ designate the downsampled probing stage, meaning tilde-accented variables are evaluated on downsampled inputs, whereas unaccented variables refer to the standard full-resolution run.
We extract the reference utilization score in the probing stage 
$\tilde{I}_{\tilde{r}}^{\tilde{t}} = \frac{1}{L}\sum_l \sum_{i \in \tilde{G} \cup P} \tilde{A}_{i, \tilde{r}}^{\tilde{t}, l}$, where vision tokens operate at a reduced spatial scale: $\tilde{R}$ (whose individual tokens are denoted by $\tilde{r}$) is obtained by VAE-encoding the downsampled RGB references, and $\tilde{G}$ represents the generative noisy tokens sampled at the reduced latent resolution.
$\tilde{t}$ is a timestep in the sparse temporal schedule, meaning the probe denoising loop executes an accelerated trajectory over a coarse time grid of only $T / D$ steps, where $D$ is the temporal reduction factor.

This probe stage shares completely the same configuration as the formal denoising stage, except that it operates on the spatially-downsampled image tokens and undergoes a temporally-downsampled DiT denoising loop (as defined by $\tilde{t}$). 
Thus, the extracted $\tilde{I}$ effectively approximates the reference utilization score $I$ of the main denoising.

\subsubsection{Reference Token Reduction}
\label{sec:method_reduction}

To better illustrate the spatial correspondence, we introduce $(w, h, n)$ as an alternative notation for the token index $r$, representing the token at position $(w, h)$ in the $n$-th reference image.
Due to the $2 \times$ spatial downsampling, each downsampled token covers a $2 \times 2$ block of full-resolution tokens.
Based on this, we apply different policies (preserve, discard, or substitute) for each $2 \times 2$ block of full-resolution reference tokens $R_{(w,h,n)}^{t, 0}$, guided by the importance score of its corresponding downsampled reference token $\tilde{I}_{(\tilde{w},\tilde{h},n)}^{\tilde{t}}$:
(1)~\textbf{Keep High-importance blocks (Full-Res Tier)} keep all 4 full-resolution tokens intact ($4 \to 4$), preserving fine-grained details.
(2)~\textbf{Discard Low-importance blocks (Discard Tier)} are completely discarded ($4 \to 0$), slashing the computational overhead of irrelevant background noise.
(3)~\textbf{Substitute Mid-importance blocks (Low-Res Tier)}: Since a simple binary strategy (keeping high-importance blocks and discarding all others) only preserves a few highly salient regions and strips away excessive global context, we introduce an intermediate level. We replace mid-importance blocks with their corresponding downsampled tokens instead of fully discarding them. This captures global context at just $1/4$ the cost.

The resulting reduced sequence $\mathcal{R}_{\text{reduced}}^{t, 0}$ is then fed into the DiT for the formal denoising stage.
This tri-tier reduction can be formulated as:
\begin{equation}
\begin{aligned}
\mathcal{R}_{\text{reduced}}^{t, 0} &= \underbrace{\big\{ R_{(w,h,n)}^{t, 0} \mid \tilde{I}_{(\tilde{w},\tilde{h},n)}^{\tilde{t}} > \tau_1 \big\}}_{\text{Preserve high-importance tokens}} 
\cup \underbrace{\big\{ \tilde{R}_{(\tilde{w},\tilde{h},n)}^{t, 0} \mid \tau_2 \le \tilde{I}_{(\tilde{w},\tilde{h},n)}^{\tilde{t}} \le \tau_1 \big\}}_{\text{Substitute mid-importance tokens}},
\end{aligned}
\end{equation}
where
$
\begin{cases} 
\tilde{w} = \lfloor w / 2 \rfloor, \ \tilde{h} = \lfloor h / 2 \rfloor \\ 
\tilde{t} = \lfloor t / D \rfloor \cdot D 
\end{cases},
\qquad
\tilde{I}_{(\tilde{w},\tilde{h},n)}^{\tilde{t}} = \frac{1}{L}\sum_l \sum_{i \in \tilde{G} \cup P} \tilde{A}_{i, (\tilde{w},\tilde{h},n)}^{\tilde{t}, l}
$.

The core rationale behind the probing stage is to approximate the main denoising behavior with minimal overhead:
Limiting the spatial downsampling to only $2\times$ ensures the probing denoising closely approximates the behaviour of main denoising.
Meanwhile, this $2\times$ downsampling cuts the 2D token count to $1/4$.
Considering attention only, the $\mathcal{O}(N^2)$ complexity reduces its cost by $16\times$; with at least $2\times$ temporal downsampling, the probe is about $\frac{1}{32}$ of the main denoising, making its overhead negligible relative to the acceleration (rigorous derivation in Appendix~\ref{sec:theo_flops}).

Unlike methods that introduce external models, the self-probing design naturally ensures optimal alignment with the generator's actual demands while avoiding the introduction of any extra parameters.
The spatial-temporal downsampling ensures that the probe introduces only minor overhead, which is significantly outweighed by the acceleration achieved through token reduction. 
This early reduction slashes FLOPs and peak VRAM while remaining orthogonal and compatible with existing in-model acceleration techniques, such as feature caching.

\section{Experiment}

\subsection{Setup}

We evaluate on three base models:
Flux.2-Klein-4B, Qwen-Image-Edit-2511,
and
HiDream-O1-Image.
See details in the Appendix~\ref{sec:suppl_models}.
We evaluate our method on two datasets: MultiBanana~\cite{MB} and Macro~\cite{macro},
which both score the generation quality from multiple dimensions.
For efficiency, we report end-to-end latency, peak activation VRAM, and speedup over the corresponding base model.
Details, including the full names of these dimensions and sub-tasks, are in the appendix~\ref{sec:suppl_data}.

\subsection{Main results}

\begin{figure*}[ht!]
  \begin{center}
    \includegraphics[width=  0.99  \textwidth]{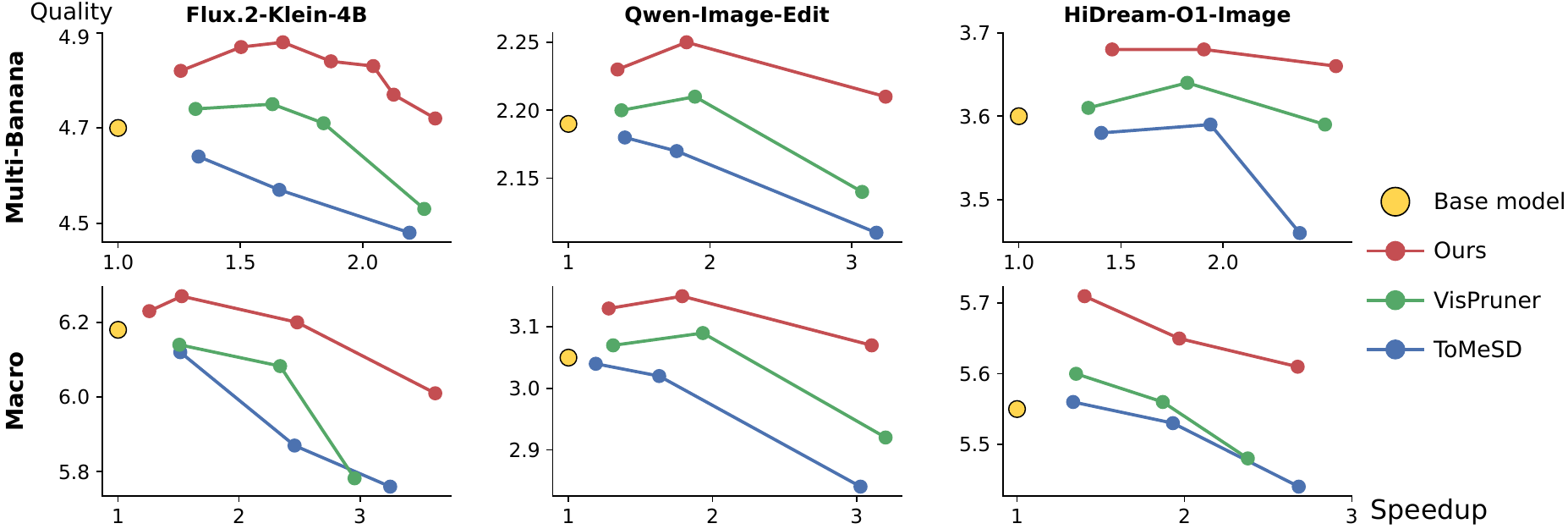}
  \end{center}
\caption{
Average quality versus speedup across three base models and two benchmarks.
}
\label{fig:pareto_as_mainComp}
\end{figure*}

\begin{table*}[ht]
\centering
\setlength{\tabcolsep}{2pt}
\resizebox{\linewidth}{!}{%
\begin{tabular}{lccccccccccccccccc}
\toprule
\multicolumn{1}{c}{\multirow{3}{*}[-6pt]{\textbf{Method}}} & \multicolumn{9}{c}{\textbf{MultiBanana}} & \multicolumn{8}{c}{\textbf{MACRO}} \\
\cmidrule(lr){2-10}\cmidrule(lr){11-18}
 & \multicolumn{6}{c}{\textbf{Quality}$\uparrow$} & \multicolumn{3}{c}{\textbf{Efficiency}} & \multicolumn{5}{c}{\textbf{Quality}$\uparrow$} & \multicolumn{3}{c}{\textbf{Efficiency}} \\
\cmidrule(lr){2-7}\cmidrule(lr){8-10}\cmidrule(lr){11-15}\cmidrule(lr){16-18}
 & \shortstack{\footnotesize Inst.\\[-3pt]\footnotesize Align} & \shortstack{\footnotesize Ref.\\[-3pt]\footnotesize Cons.} & \shortstack{\footnotesize B-S\\[-3pt]\footnotesize Match.} & \shortstack{\footnotesize Phys.\\[-3pt]\footnotesize Real.} & \shortstack{\footnotesize Visu.\\[-3pt]\footnotesize Qual.} & \raisebox{3.5pt}{\textbf{Avg}} & \shortstack{\footnotesize Latency\\[-3pt]\footnotesize (s)$\downarrow$} & \raisebox{3.5pt}{Speed$\uparrow$} & \shortstack{\footnotesize Act. VRAM\\[-3pt]\footnotesize (GB)$\downarrow$} & \shortstack{\footnotesize Cust.} & \shortstack{\footnotesize Illus.} & \shortstack{\footnotesize Spat.} & \shortstack{\footnotesize Temp.} & \raisebox{3.5pt}{\textbf{Avg}} & \shortstack{\footnotesize Latency\\[-3pt]\footnotesize (s)$\downarrow$} & \raisebox{3.5pt}{Speed$\uparrow$} & \shortstack{\footnotesize Act. VRAM\\[-3pt]\footnotesize (GB)$\downarrow$} \\
\midrule
Baseline & 4.57 & 4.86 & 3.96 & 4.40 & 5.72 & 4.70 & 5.74 & 1.00$\times$ & 15.3 & 7.49 & 6.17 & 3.83 & 7.23 & 6.18 & 7.93 & 1.00$\times$ & 24.7 \\
\midrule
\multicolumn{18}{l}{\textit{Fast}} \\
ToMeSD & 4.45 & 4.51 & 3.72 & 4.17 & 5.55 & 4.48 & 2.62 & 2.19$\times$ & 11.2 & 7.10 & 5.82 & 3.45 & 7.10 & 5.87 & 3.23 & 2.46$\times$ & 17.4 \\
VisPruner & 4.54 & 4.53 & 3.72 & 4.24 & 5.62 & 4.53 & 2.55 & 2.25$\times$ & 8.7 & 7.27 & 5.99 & 3.87 & 7.20 & 6.08 & 3.39 & 2.34$\times$ & 13.7 \\
Ours & \textbf{4.63} & \textbf{4.85} & \textbf{3.97} & \textbf{4.38} & \textbf{5.77} & \textbf{4.72} & 2.50 & 2.30$\times$ & 7.3 & \textbf{7.50} & \textbf{6.15} & \textbf{3.88} & \textbf{7.27} & \textbf{6.20} & 3.20 & 2.48$\times$ & 11.5 \\
\midrule
\multicolumn{18}{l}{\textit{Slow}} \\
ToMeSD & 4.48 & 4.53 & 3.82 & 4.38 & 5.62 & 4.57 & 3.46 & 1.66$\times$ & 10.7 & 7.35 & 6.18 & 3.70 & 7.25 & 6.12 & 5.24 & 1.51$\times$ & 17.4 \\
VisPruner & 4.68 & 4.85 & 3.90 & 4.44 & 5.88 & 4.75 & 3.52 & 1.63$\times$ & 10.2 & 7.37 & 6.03 & 3.91 & 7.25 & 6.14 & 5.27 & 1.51$\times$ & 17.2 \\
Ours & \textbf{4.86} & \textbf{5.06} & \textbf{4.03} & \textbf{4.54} & \textbf{5.92} & \textbf{4.88} & 3.43 & 1.67$\times$ & 9.8 & \textbf{7.55} & \textbf{6.20} & \textbf{4.01} & \textbf{7.32} & \textbf{6.27} & 5.20 & 1.53$\times$ & 16.4 \\
\bottomrule
\end{tabular}}
\caption{Comparison of efficiency and quality on MultiBanana and MACRO (Flux.2-Klein-4B).}
\label{tab:mainB}
\end{table*}

\begin{figure*}[ht]
  \begin{center}
    \includegraphics[width=  1  \textwidth]{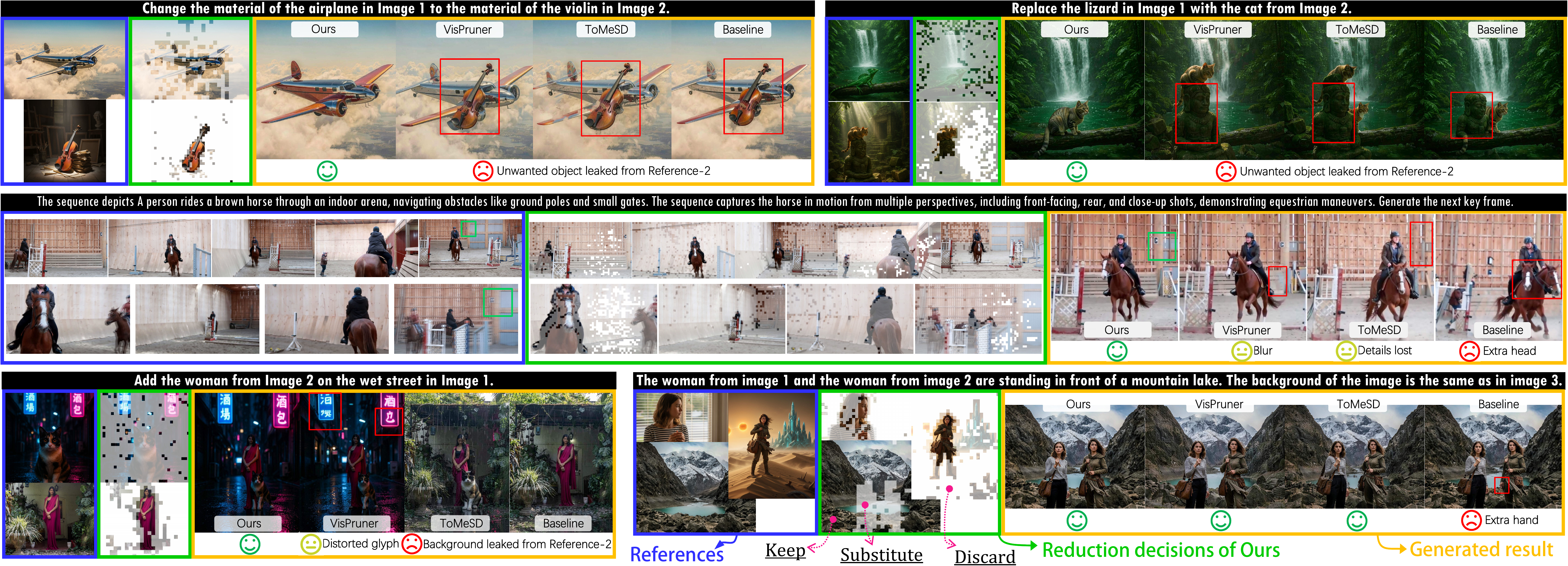}
  \end{center}
  \caption{
Visual comparison of generation quality across diverse tasks from MultiBanana and Macro datasets (all methods are under the Slow level in Table~\ref{tab:mainB} with comparable speedups).
Best viewed when zoomed in.
}
\label{fig:comp}
\end{figure*}

Fig.~\ref{fig:pareto_as_mainComp} presents the main comparison across three base models and two benchmarks, against the original base model (\textbf{Baseline}), ToMeSD~\cite{tomesd} (applied before DiT as a pre-model method), and VisPruner~\cite{visPruner}.
Table~\ref{tab:mainB} reports the fine-grained scores of this comparison on Flux.2-Klein-4B.
\textbf{Ours(fast)} accelerates inference and lowers peak VRAM, while matching or surpassing baseline quality;
\textbf{Ours(slow)} further improves all quality metrics, at a smaller speedup.
VisPruner can improve quality over the baseline at some speedup, 
while its quality stays below ours because it identifies unimportant tokens less accurately.
Fig.~\ref{fig:comp} shows visual results.

\subsection{Orthogonality with in-model methods}

We compare with in-model acceleration methods including TaylorSeer~\cite{taylorseer}, RepreShift~\cite{repreShift}, and TransPrune~\cite{transPrune} in Table~\ref{tab:inmodel}.
As demonstrated in Table~\ref{tab:inmodel}, integrating our method with existing in-model methods yields compounded speedups.
Detailed discussion of our orthogonality with other existing methods can be found in Appendix~\ref{sec:suppl_orth}.

\begin{table*}[ht]
\centering
\setlength{\tabcolsep}{2pt}
\resizebox{\linewidth}{!}{%
\begin{tabular}{lccccccccccccccccc}
\toprule
\multicolumn{1}{c}{\multirow{3}{*}[-6pt]{\textbf{Method}}} & \multicolumn{9}{c}{\textbf{MultiBanana}} & \multicolumn{8}{c}{\textbf{MACRO}} \\
\cmidrule(lr){2-10}\cmidrule(lr){11-18}
 & \multicolumn{6}{c}{\textbf{Quality}$\uparrow$} & \multicolumn{3}{c}{\textbf{Efficiency}} & \multicolumn{5}{c}{\textbf{Quality}$\uparrow$} & \multicolumn{3}{c}{\textbf{Efficiency}} \\
\cmidrule(lr){2-7}\cmidrule(lr){8-10}\cmidrule(lr){11-15}\cmidrule(lr){16-18}
 & \shortstack{\footnotesize Inst.\\[-3pt]\footnotesize Align} & \shortstack{\footnotesize Ref.\\[-3pt]\footnotesize Cons.} & \shortstack{\footnotesize B-S\\[-3pt]\footnotesize Match.} & \shortstack{\footnotesize Phys.\\[-3pt]\footnotesize Real.} & \shortstack{\footnotesize Visu.\\[-3pt]\footnotesize Qual.} & \raisebox{3.5pt}{\textbf{Avg}} & \shortstack{\footnotesize Latency\\[-3pt]\footnotesize (s)$\downarrow$} & \raisebox{3.5pt}{Speed$\uparrow$} & \shortstack{\footnotesize Act. VRAM\\[-3pt]\footnotesize (GB)$\downarrow$} & \shortstack{\footnotesize Cust.} & \shortstack{\footnotesize Illus.} & \shortstack{\footnotesize Spat.} & \shortstack{\footnotesize Temp.} & \raisebox{3.5pt}{\textbf{Avg}} & \shortstack{\footnotesize Latency\\[-3pt]\footnotesize (s)$\downarrow$} & \raisebox{3.5pt}{Speed$\uparrow$} & \shortstack{\footnotesize Act. VRAM\\[-3pt]\footnotesize (GB)$\downarrow$} \\
\midrule
Baseline & 4.57 & 4.86 & 3.96 & 4.40 & 5.72 & 4.70 & 5.7 & 1.00$\times$ & 15.3 & 7.49 & 6.17 & 3.83 & 7.23 & 6.18 & 7.9 & 1.00$\times$ & 24.7 \\
Ours &  4.63 &  4.85 &  3.97 &  4.38 &  5.77 &  4.72 & 2.5 & 2.30$\times$ & 7.3 &  7.50 &  6.15 &  3.88 &  7.27 &  6.20 & 3.2 & 2.48$\times$ & 11.5 \\
\hdashline
RepreShift & 4.51 & 4.69 & 3.93 & 4.44 & 5.78 & 4.67 & 3.5 & 1.63$\times$ & 11.3 & 7.45 & 6.12 & 3.83 & 7.11 & 6.13 & 4.1 & 1.93$\times$ & 22.1 \\
RepreShift + Ours & 4.61 & 4.66 & 3.96 & 4.40 & 5.72 & 4.67 & 2.4 & 2.38$\times$ & 8.3 & 7.44 & 6.10 & 3.85 & 7.11 & 6.12 & 3.0 & 2.65$\times$ & 15.2 \\
\hdashline
Taylorseer & 4.52 & 4.69 & 3.85 & 4.41 & 5.64 & 4.62 & 4.8 & 1.19$\times$ & 47.5 & 7.13 & 5.63 & 3.16 & 6.82 & 5.69 & 6.5 & 1.21$\times$ & 62.0 \\
Taylorseer + Ours & 4.55 & 4.72 & 3.90 & 4.44 & 5.70 & 4.66 & 3.1 & 1.85$\times$ & 34.3 & 7.21 & 5.47 & 3.26 & 6.86 & 5.70 & 4.8 & 1.65$\times$ & 45.4 \\
\hdashline
TransPrune & 4.56 & 4.85 & 3.92 & 4.43 & 5.81 & 4.71 & 3.2 & 1.77$\times$ & 14.9 & 7.52 & 6.15 & 3.87 & 7.17 & 6.18 & 3.3 & 2.40$\times$ & 23.3 \\
TransPrune + Ours & 4.58 & 4.82 & 3.94 & 4.38 & 5.77 & 4.70 & 2.1 & 2.79$\times$ & 8.4 & 7.48 & 6.14 & 3.88 & 7.13 & 6.16 & 2.5 & 3.12$\times$ & 15.6 \\
\bottomrule
\end{tabular}}
\caption{
Compatibility of Ours(fast) with existing in-model acceleration methods on FLUX.2-Klein-4B, yielding compounded inference speedups.
}
\label{tab:inmodel}
\end{table*}

\subsection{Ablation study}

\begin{table*}[ht]
\centering
\setlength{\aboverulesep}{0pt}
\setlength{\belowrulesep}{0pt}
\setlength{\tabcolsep}{4pt}
\resizebox{\linewidth}{!}{%
\begin{tabular}{l|cccccc}
\toprule
\raisebox{3.5pt}{\textbf{Method}} & \shortstack{\footnotesize Instruction\\[-3pt]\footnotesize Alignment$\uparrow$} & \shortstack{\footnotesize Reference\\[-3pt]\footnotesize Consistency$\uparrow$} & \shortstack{\footnotesize BG-Subject\\[-3pt]\footnotesize Match$\uparrow$} & \shortstack{\footnotesize Physical\\[-3pt]\footnotesize Realism$\uparrow$} & \shortstack{\footnotesize Visual\\[-3pt]\footnotesize Quality$\uparrow$} & \raisebox{3.5pt}{\textbf{Overall}$\uparrow$} \\
\midrule
Only Low-Res Tier (i.e., infer baseline on 2$\times$ downsampled images) & 4.25 & 4.10 & 3.53 & 4.12 & 5.63 & 4.33 \\
No Low-Res Tier & 4.56 & 4.67 & 3.88 & 4.37 & 5.72 & 4.64 \\
Random Reduction & 4.25 & 4.08 & 3.62 & 4.22 & 5.60 & 4.35 \\
Ours & \textbf{4.63} & \textbf{4.85} & \textbf{3.97} & \textbf{4.38} & \textbf{5.77} & \textbf{4.72} \\
\bottomrule
\end{tabular}}
\caption{Ablation study on Ours(fast) on MultiBanana (controlled to the same reduction ratio). Base model: Flux.2-Klein-4B.
}
\label{tab:abla}
\end{table*}

Table~\ref{tab:abla} validates our designs.
``All Low-Res Tier'' operates on the $2\times$ downsampled references, suffering a severe quality drop.
Removing the intermediate tier (``w/o Low-Res Tier'') causes a noticeable quality decline, confirming the necessity of capturing global context via Low-Res Tier tokens.
``Random Reduction'' also degrades quality, indicating the extracted reference utilization score is necessary for token reduction decisions.

\subsection{Analysis of Quality Improvement}

\subsubsection{Effect of Token Reduction Ratio}

\begin{wrapfigure}{r}{0.70\textwidth}
  \centering
  \includegraphics[width=\linewidth]{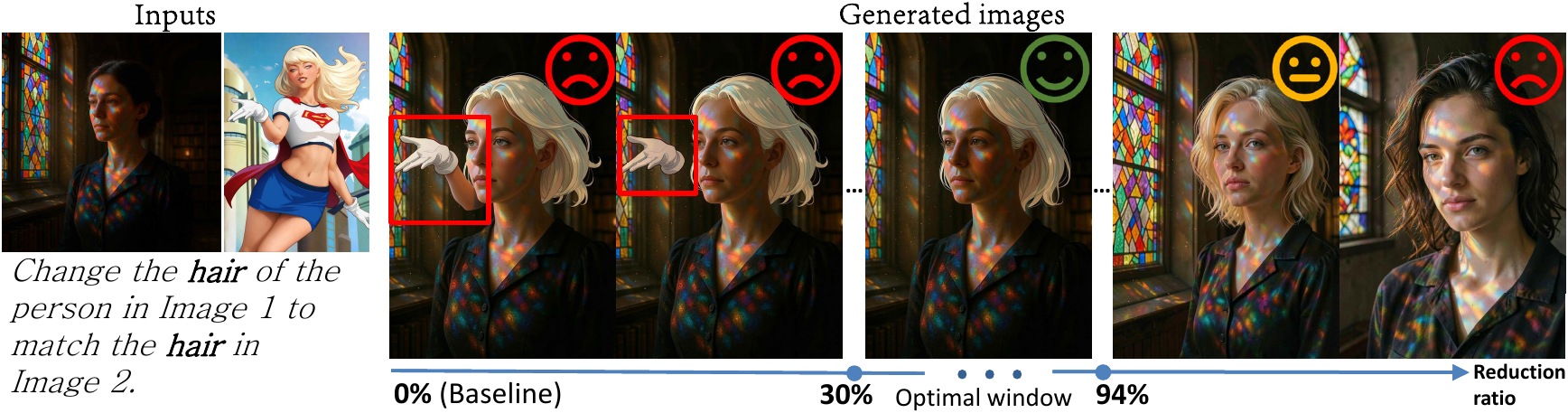}
  \caption{
  Qualitative evolution of generated images. At 0\% reduction (unpruned baseline), the model mistakenly retains visual distractors (e.g., an extra floating glove). Within a broad window (30\% to 94\%), our method successfully filters out these distractors, boosting quality. At extreme reduction levels ($>95$\%), the irreversible loss of informative reference tokens causes quality degradation.
  }
  \label{fig:reductionRatio}
\end{wrapfigure}
Fig.~\ref{fig:reductionRatio} shows how quality evolves with the reduction ratio, driven by two competing factors: (1) mitigating attention interference and dilution versus (2) losing informative tokens at extreme sparsity.
At 0\% reduction, the baseline leaks a floating glove from a salient but irrelevant region in ref~2;
Moving into a broad optimal window, Factor 1 dominates, and the model successfully filters out these visual distractors.
As the reduction ratio escalates to extreme levels ($>$95\%), Factor 2 eventually takes over. Crucial features are irreversibly pruned, resulting in degraded reference consistency.
Note that this \textbf{optimal window} forms a \textbf{wide} plateau rather than a narrow range, as demonstrated by the case in Fig.~\ref{fig:reductionRatio} and quantitative curve in Fig.~\ref{fig:pareto_as_mainComp}.

\subsubsection{Effective Attention Ratio}
\label{sec:EAR}

Our method can effectively mitigate attention dilution and attention interference, thus making the model allocate its limited attention capacity to genuinely useful reference tokens. Here, we explicitly validate whether the model's attention is more focused on meaningful reference regions, by evaluating the Effective Attention Ratio (EAR).

\begin{table*}[ht]
\centering
\setlength{\aboverulesep}{0pt}
\setlength{\belowrulesep}{0pt}
\setlength{\tabcolsep}{3pt}
\resizebox{\linewidth}{!}{%
\begin{tabular}{l|c|cccccc}
\toprule
\raisebox{3.5pt}{\textbf{Layers used for $I_r$ Extraction}} & \raisebox{3.5pt}{\textbf{Effective Attention Ratio}$\uparrow$} & \shortstack{\footnotesize Instruction\\[-3pt]\footnotesize Alignment$\uparrow$} & \shortstack{\footnotesize Reference\\[-3pt]\footnotesize Consistency$\uparrow$} & \shortstack{\footnotesize BG-Subject\\[-3pt]\footnotesize Match$\uparrow$} & \shortstack{\footnotesize Physical\\[-3pt]\footnotesize Realism$\uparrow$} & \shortstack{\footnotesize Visual\\[-3pt]\footnotesize Quality$\uparrow$} & \raisebox{3.5pt}{\textbf{Overall}$\uparrow$} \\
\midrule
Baseline (No Reduction) & 43.6 & 4.57 & 4.86 & 3.96 & 4.40 & 5.72 & 4.70 \\
\hdashline
Top-1 layer & 51.7 & 4.81 & 5.08 & 3.92 & 4.42 & 5.85 & 4.82 \\
Top $20\%$ of layers & \textbf{52.9} & \textbf{4.89} & \textbf{5.13} & 4.01 & 4.47 & \textbf{5.93} & \textbf{4.89} \\
All layers & 52.7 & 4.86 & 5.06 & \textbf{4.03} & \textbf{4.54} & 5.92 & 4.88 \\
\bottomrule
\end{tabular}}
\caption{Effective Attention Ratio (EAR) on all samples from the object composition task of MultiBanana.
Baseline: Flux.2-Klein-4B. 
All variants use the same reduction ratio (Slow level).
}
\label{tab:ear}
\end{table*}

To calculate EAR, an objective ground-truth (GT) importance map is required. While general reference-based tasks lack such GT importance, the object composition task in MultiBanana~\cite{MB} inherently aims to composite the salient object in each reference image. Thus, the salient object region (extracted via BiRefNet~\cite{birefnet}) naturally serves as the GT important area. 
A higher ratio indicates that the model allocates more attention to genuinely useful regions.
We evaluate four variants: (1)~\textit{Baseline (No Reduction)}, the unpruned reference sequence; (2)~\textit{Top-1 layer}, using the utilization score from the best-aligned single layer; (3)~\textit{Top 20\% of layers}, using the mean over the top 20\% best-correlated layers; and (4)~\textit{All layers}, using the mean across all layers. Details are in Appendix~\ref{sec:suppl_ear}.

Table~\ref{tab:ear} shows two gains. Token reduction with \textit{Top-1 layer} markedly raises EAR, validating that our token reduction makes the model allocate more attention to genuinely useful reference regions. \textit{Top 20\% of layers} and \textit{All layers} further slightly raise EAR over \textit{Top-1 layer}, validating that averaging across layers further mitigates interference by identifying unimportant tokens more accurately.

\textit{All layers} is comparable to \textit{Top 20\% of layers}, but does not require model-specific layer selection. We therefore adopt averaging across all layers as the default for generalization across base models.

\section{Conclusions}
In this work, we propose an early token reduction framework via downsampled self-probing to address the token explosion in reference-based diffusion models. By utilizing the model itself to cheaply extract token importance, we perform a tri-tier reduction before the main denoising stage. Experiments show that our training-free method reduces latency, mitigate peak VRAM, seamlessly compounds with in-model acceleration techniques, and improves generation quality.

\bibliographystyle{iclr2027_conference}
\bibliography{refs}

@inproceedings{lin2025ctrl,
  author        = {Han Lin and Jaemin Cho and Abhay Zala and Mohit Bansal},
  title         = {Ctrl-Adapter: An Efficient and Versatile Framework for Adapting Diverse Controls to Any Diffusion Model},
  booktitle     = {International Conference on Learning Representations (ICLR)},
  year          = {2025},
  url           = {https://arxiv.org/abs/2404.09967},
  archiveprefix = {arXiv},
  eprint        = {2404.09967},
  primaryclass  = {cs.CV},
}

@inproceedings{gong2026relationadapter,
  author        = {Yan Gong and Yiren Song and Yicheng Li and Chenglin Li and Yin Zhang},
  title         = {RelationAdapter: Learning and Transferring Visual Relation with Diffusion Transformers},
  booktitle     = {Advances in Neural Information Processing Systems (NeurIPS)},
  year          = {2025},
  url           = {https://arxiv.org/abs/2506.02528},
  archiveprefix = {arXiv},
  eprint        = {2506.02528},
  primaryclass  = {cs.CV},
}

@misc{unibiotransfer,
  author        = {Caiyi Sun and Yujing Sun and Xiangyu Li and Yuhang Zheng and Yiming Ren and Jiamin Wang and Yuexin Ma and Siu-Ming Yiu},
  title         = {UniBioTransfer: A Unified Framework for Multiple Biometrics Transfer},
  year          = {2026},
  url           = {https://arxiv.org/abs/2603.19637},
  archiveprefix = {arXiv},
  eprint        = {2603.19637},
  howpublished  = {arXiv preprint arXiv:2603.19637},
  primaryclass  = {cs.CV},
}

@misc{Durian,
  author        = {Hyunsoo Cha and Byungjun Kim and Hanbyul Joo},
  title         = {Durian: Dual Reference Image-Guided Portrait Animation with Attribute Transfer},
  year          = {2025},
  url           = {https://arxiv.org/abs/2509.04434},
  archiveprefix = {arXiv},
  eprint        = {2509.04434},
  howpublished  = {arXiv preprint arXiv:2509.04434},
  primaryclass  = {cs.CV},
}

@misc{flux2,
    author={{Black Forest Labs}},
    title={{FLUX.2: Frontier Visual Intelligence}},
    year={2025},
    howpublished={\url{https://bfl.ai/blog/flux-2}},
}

@misc{qwenImage,
  author        = {Chenfei Wu and Jiahao Li and Jingren Zhou and Junyang Lin and Kaiyuan Gao and Kun Yan and Sheng-ming Yin and Shuai Bai and Xiao Xu and Yilei Chen and Yuxiang Chen and Zecheng Tang and Zekai Zhang and Zhengyi Wang and An Yang and Bowen Yu and Chen Cheng and Dayiheng Liu and Deqing Li and Hang Zhang and Hao Meng and Hu Wei and Jingyuan Ni and Kai Chen and Kuan Cao and Liang Peng and Lin Qu and Minggang Wu and Peng Wang and Shuting Yu and Tingkun Wen and Wensen Feng and Xiaoxiao Xu and Yi Wang and Yichang Zhang and Yongqiang Zhu and Yujia Wu and Yuxuan Cai and Zenan Liu},
  title         = {Qwen-Image Technical Report},
  year          = {2025},
  url           = {https://arxiv.org/abs/2508.02324},
  archiveprefix = {arXiv},
  eprint        = {2508.02324},
  howpublished  = {arXiv preprint arXiv:2508.02324},
  primaryclass  = {cs.CV},
}

@article{firered,
  title={FireRed-Image-Edit-1.0 Technical Report}, 
  author={{Super Intelligence Team}},
  year={2026},
  archivePrefix={arXiv},
  primaryClass={cs.CV},
  url={https://arxiv.org/abs/2602.13344}, 
}

@misc{hiDream,
  author        = {Qi Cai and Jingwen Chen and Chengmin Gao and Zijian Gong and Yehao Li and Yingwei Pan and Yi Peng and Zhaofan Qiu and Kai Yu and Yiheng Zhang and Hao Ai and Siying Bai and Yang Chen and Zhihui Chen and Fengbin Gao and Ying Guo and Dong Li and Zhen Shen and Leilei Shi and Jing Wang and Siyu Wang and Yimeng Wang and Rui Zheng and Ting Yao and Tao Mei},
  title         = {HiDream-O1-Image: A Natively Unified Image Generative Foundation Model with Pixel-level Unified Transformer},
  year          = {2026},
  url           = {https://arxiv.org/abs/2605.11061},
  archiveprefix = {arXiv},
  eprint        = {2605.11061},
  howpublished  = {arXiv preprint arXiv:2605.11061},
  primaryclass  = {cs.CV},
}

@inproceedings{tomesd,
  author        = {Bolya, Daniel and Hoffman, Judy},
  title         = {Token Merging for Fast Stable Diffusion},
  booktitle     = {IEEE Conference on Computer Vision and Pattern Recognition -- Workshops (CVPRW)},
  year          = {2023},
  url           = {https://arxiv.org/abs/2303.17604},
  archiveprefix = {arXiv},
  eprint        = {2303.17604},
  primaryclass  = {cs.CV},
}

@inproceedings{visPruner,
  author       = {Qizhe Zhang and
                  Aosong Cheng and
                  Ming Lu and
                  Renrui Zhang and
                  Zhiyong Zhuo and
                  Jiajun Cao and
                  Shaobo Guo and
                  Qi She and
                  Shanghang Zhang},
  title        = {Beyond Text-Visual Attention: Exploiting Visual Cues for Effective
                  Token Pruning in VLMs},
  booktitle    = {{IEEE/CVF} International Conference on Computer Vision, {ICCV} 2025,
                  Honolulu, HI, USA, October 19-25, 2025},
  pages        = {20857--20867},
  publisher    = {{IEEE}},
  year         = {2025},
  url          = {https://doi.org/10.1109/ICCV51701.2025.01939},
  doi          = {10.1109/ICCV51701.2025.01939},
  bibsource    = {dblp computer science bibliography, https://dblp.org}
}

@inproceedings{fastv,
  title={An image is worth 1/2 tokens after layer 2: Plug-and-play inference acceleration for large vision-language models},
  author={Chen, Liang and Zhao, Haozhe and Liu, Tianyu and Bai, Shuai and Lin, Junyang and Zhou, Chang and Chang, Baobao},
  booktitle={European Conference on Computer Vision},
  pages={19--35},
  year={2024},
  organization={Springer}
}

@article{hired,
  title={Hired: Attention-guided token dropping for efficient inference of high-resolution vision-language models},
  author={Arif, Kazi Hasan Ibn and Yoon, JinYi and Nikolopoulos, Dimitrios S and Vandierendonck, Hans and John, Deepu and Ji, Bo},
  booktitle={Proceedings of the AAAI Conference on Artificial Intelligence},
  volume={39},
  number={2},
  pages={1773--1781},
  year={2025}
}

@inproceedings{prumerge,
  title={Llava-prumerge: Adaptive token reduction for efficient large multimodal models},
  author={Shang, Yuzhang and Cai, Mu and Xu, Bingxin and Lee, Yong Jae and Yan, Yan},
  booktitle={2025 IEEE/CVF International Conference on Computer Vision (ICCV)},
  pages={22857--22867},
  year={2025},
  organization={IEEE}
}

@inproceedings{visionzip,
  title={Visionzip: Longer is better but not necessary in vision language models},
  author={Yang, Senqiao and Chen, Yukang and Tian, Zhuotao and Wang, Chengyao and Li, Jingyao and Yu, Bei and Jia, Jiaya},
  booktitle={2025 IEEE/CVF Conference on Computer Vision and Pattern Recognition (CVPR)},
  pages={19792--19802},
  year={2025},
  organization={IEEE}
}

@article{flexselect,
  title={Flexselect: Flexible token selection for efficient long video understanding},
  author={Lu, Yu and Wang, Tianyi and Rao, Fengyun and Yang, Yi and Zhu, Linchao and others},
  journal={Advances in Neural Information Processing Systems},
  volume={38},
  pages={102751--102777},
  year={2026}
}

@article{topi,
  title={Token Pruning for In-Context Generation in Diffusion Transformers},
  author={Lin, Junqing and Zheng, Xingyu and Cheng, Pei and Fu, Bin and Sun, Jingwei and Sun, Guangzhong},
  journal={arXiv preprint arXiv:2602.01609},
  year={2026}
}

@inproceedings{sdtm,
  author        = {Fang, Haipeng and Tang, Sheng and Cao, Juan and Zhang, Enshuo and Tang, Fan and Lee, Tong-Yee},
  title         = {Attend to Not Attended: Structure-then-Detail Token Merging for Post-training DiT Acceleration},
  booktitle     = {IEEE/CVF Conference on Computer Vision and Pattern Recognition (CVPR)},
  year          = {2025},
  url           = {https://arxiv.org/abs/2505.11707},
  archiveprefix = {arXiv},
  eprint        = {2505.11707},
  primaryclass  = {cs.CV},
}

@misc{CoReDiT,
  author        = {Zhuojin Li and Hsin-Pai Cheng and Hong Cai and Shizhong Han and Fatih Porikli},
  title         = {CoReDiT: Spatial Coherence-Guided Token Pruning and Reconstruction for Efficient Diffusion Transformers},
  year          = {2026},
  url           = {https://arxiv.org/abs/2605.14191},
  archiveprefix = {arXiv},
  eprint        = {2605.14191},
  howpublished  = {arXiv preprint arXiv:2605.14191},
  primaryclass  = {cs.CV},
}

@inproceedings{diffsparse,
  title={Diffsparse: Accelerating diffusion transformers with learned token sparsity},
  author={Zhu, Haowei and Liu, Ji and Liu, Ziqiong and Li, Dong and Yong, Jun-Hai and Wang, Bin and Barsoum, Emad},
  booktitle={International Conference on Learning Representations},
  volume={2026},
  pages={57438--57456},
  year={2026}
}

@inproceedings{deepcache,
  author        = {Ma, Xinyin and Fang, Gongfan and Wang, Xinchao},
  title         = {DeepCache: Accelerating Diffusion Models for Free},
  booktitle     = {IEEE/CVF Conference on Computer Vision and Pattern Recognition (CVPR)},
  year          = {2024},
  url           = {https://arxiv.org/abs/2312.00858},
  archiveprefix = {arXiv},
  eprint        = {2312.00858},
  primaryclass  = {cs.CV},
}

@article{delta_dit,
  title={{\(\Delta\)}-DiT: {A} Training-Free Acceleration Method Tailored for Diffusion Transformers},
  author={Chen, Pengtao and Shen, Mingzhu and Ye, Peng and Cao, Jianjian and Tu, Chongjun and Bouganis, Christos-Savvas and Zhao, Yiren and Chen, Tao},
  journal={arXiv preprint arXiv:2406.01125},
  year={2024}
}

@inproceedings{toca,
 author = {Zou, Chang and Liu, Xuyang and Liu, Ting and Huang, Siteng and Zhang, Linfeng},
 booktitle = {International Conference on Learning Representations},
 editor = {Y. Yue and A. Garg and N. Peng and F. Sha and R. Yu},
 pages = {75467--75488},
 title = {Accelerating Diffusion Transformers with Token-wise Feature Caching},
 url = {https://proceedings.iclr.cc/paper_files/paper/2025/file/bbe024e0517fe12ac3a8d388b19ff9fe-Paper-Conference.pdf},
 volume = {2025},
 year = {2025}
}

@article{teacache,
  title={Timestep Embedding Tells: It's Time to Cache for Video Diffusion Model},
  author={Liu, Feng and Zhang, Shiwei and Wang, Xiaofeng and Wei, Yujie and Qiu, Haonan and Zhao, Yuzhong and Zhang, Yingya and Ye, Qixiang and Wan, Fang},
  journal={arXiv preprint arXiv:2411.19108},
  year={2024}
}

@article{taylorseer,
  title={From reusing to forecasting: Accelerating diffusion models with taylorseers},
  author={Liu, Jiacheng and Zou, Chang and Lyu, Yuanhuiyi and Chen, Junjie and Zhang, Linfeng},
  booktitle={2025 IEEE/CVF International Conference on Computer Vision (ICCV)},
  pages={15853--15863},
  year={2025},
  organization={IEEE}
}

@inproceedings{l2c,
  author        = {Ma, Xinyin and Fang, Gongfan and Mi, Michael Bi and Wang, Xinchao},
  title         = {Learning-to-Cache: Accelerating Diffusion Transformer via Layer Caching},
  booktitle     = {Advances in Neural Information Processing Systems (NeurIPS)},
  year          = {2024},
  url           = {https://arxiv.org/abs/2406.01733},
  archiveprefix = {arXiv},
  collection    = {NeurIPS 2024},
  eprint        = {2406.01733},
  primaryclass  = {cs.LG},
}

@inproceedings{spotedit,
  title={SpotEdit: Selective Region Editing in Diffusion Transformers},
  author={Qin, Zhibin and Tan, Zhenxiong and Wang, Zeqing and Liu, Songhua and Wang, Xinchao},
  booktitle={CVPR},
  year={2026}
}

@article{evctrl,
  title={EVCtrl: Efficient Control Adapter for Visual Generation},
  author={Yang, Zixiang and Ma, Yue and Zhang, Yinhan and Mo, Shanhui and Liu, Dongrui and Zhang, Linfeng},
  journal={arXiv preprint arXiv:2508.10963},
  year={2025}
}

@article{regionE,
  title={RegionE: Adaptive Region-Aware Generation for Efficient Image Editing},
  author={Chen, Pengtao and Zeng, Xianfang and Zhao, Maosen and Shen, Mingzhu and Ye, Peng and Xiang, Bangyin and Wang, Zhibo and Cheng, Wei and Yu, Gang and Chen, Tao},
  journal={arXiv preprint arXiv:2510.25590},
  year={2025}
}

@inproceedings{deepseek_NSA,
  title={Native Sparse Attention: Hardware-Aligned and Natively Trainable Sparse Attention},
  author={Yuan, Jingyang and Gao, Huazuo and Dai, Damai and Luo, Junyu and Zhao, Liang and Zhang, Zhengyan and Xie, Zhenda and Wei, Yuxing and Wang, Lean and Xiao, Zhiping and others},
  booktitle={Proceedings of the 63rd Annual Meeting of the Association for Computational Linguistics},
  pages={23078--23097},
  year={2025}
}

@inproceedings{zhou2026trainable_nlog,
  title={Trainable Log-Linear Sparse Attention for Efficient Diffusion Transformers},
  author={Zhou, Yifan and Xiao, Zeqi and Wei, Tianyi and Yang, Shuai and Pan, Xingang},
  booktitle={CVPR},
  pages={9424--9433},
  year={2026}
}

@article{linfusion,
  title     = {LinFusion: 1 GPU, 1 Minute, 16K Image},
  author    = {Liu, Songhua and Yu, Weihao and Tan, Zhenxiong and Wang, Xinchao},
  year      = {2024},
  eprint    = {2409.02097},
  archivePrefix={arXiv},
  primaryClass={cs.CV}
}

@misc{sana,
  title={Sana: Efficient High-Resolution Image Synthesis with Linear Diffusion Transformer},
  author={Enze Xie and Junsong Chen and Junyu Chen and Han Cai and Haotian Tang and Yujun Lin and Zhekai Zhang and Muyang Li and Ligeng Zhu and Yao Lu and Song Han},
  year={2024},
  eprint={2410.10629},
  archivePrefix={arXiv},
  primaryClass={cs.CV},
  url={https://arxiv.org/abs/2410.10629},
}

@article{survey_mllm_token_compress_2026,
  title={Towards Efficient Multimodal Large Language Models: A Survey on Token Compression},
  author={Yao, Linli and Xing, Long and Shi, Yang and Li, Sida and Liu, Yuanxin and Dong, Yuhao and Zhang, Yi-Fan and Li, Lei and Dong, Qingxiu and Dong, Xiaoyi and others},
  year={2026},
  publisher={TechRxiv}
}

@article{survey_token_reduction_position,
  title={Token Reduction Should Go Beyond Efficiency in Generative Models -- From Vision, Language to Multimodality},
  author={Kong, Zhenglun and Li, Yize and Zeng, Fanhu and Xin, Lei and Messica, Shvat and Lin, Xue and Zhao, Pu and Kellis, Manolis and Tang, Hao and Zitnik, Marinka},
  journal={arXiv preprint arXiv:2505.18227},
  year={2025}
}

@article{survey_mllm_compress_2026,
  title={A Survey of Token Compression for Efficient Multimodal Large Language Models},
  author={Shao, Kele and Tao, Keda and Zhang, Kejia and Feng, Sicheng and Cai, Mu and Shang, Yuzhang and You, Haoxuan and Qin, Can and Sui, Yang and Wang, Huan},
  journal={arXiv preprint arXiv:2507.20198},
  year={2026}
}

@misc{lcm,
      title={Latent Consistency Models: Synthesizing High-Resolution Images with Few-Step Inference}, 
      author={Simian Luo and Yiqin Tan and Longbo Huang and Jian Li and Hang Zhao},
      year={2023},
      eprint={2310.04378},
      archivePrefix={arXiv},
      primaryClass={cs.CV}
}

@inproceedings{add,
  author        = {Sauer, Axel and Lorenz, Dominik and Blattmann, Andreas and Rombach, Robin},
  title         = {Adversarial Diffusion Distillation},
  booktitle     = {European Conference on Computer Vision (ECCV)},
  year          = {2024},
  url           = {https://arxiv.org/abs/2311.17042},
  archiveprefix = {arXiv},
  eprint        = {2311.17042},
  primaryclass  = {cs.CV},
}

@inproceedings{qdiff,
  author        = {Li, Xiuyu and Liu, Yijiang and Lian, Long and Yang, Huanrui and Dong, Zhen and Kang, Daniel and Zhang, Shanghang and Keutzer, Kurt},
  title         = {Q-Diffusion: Quantizing Diffusion Models},
  booktitle     = {IEEE/CVF International Conference on Computer Vision (ICCV)},
  year          = {2023},
  url           = {https://arxiv.org/abs/2302.04304},
  archiveprefix = {arXiv},
  eprint        = {2302.04304},
  primaryclass  = {cs.CV},
}

@inproceedings{qdit,
  author        = {Chen, Lei and Meng, Yuan and Tang, Chen and Ma, Xinzhu and Jiang, Jingyan and Wang, Xin and Wang, Zhi and Zhu, Wenwu},
  title         = {Q-DiT: Accurate Post-Training Quantization for Diffusion Transformers},
  booktitle     = {IEEE/CVF Conference on Computer Vision and Pattern Recognition (CVPR)},
  year          = {2025},
  url           = {https://arxiv.org/abs/2406.17343},
  archiveprefix = {arXiv},
  eprint        = {2406.17343},
  primaryclass  = {cs.CV},
}

@misc{DaTo,
  author        = {Evelyn Zhang and Bang Xiao and Jiayi Tang and Qianli Ma and Chang Zou and Xuefei Ning and Xuming Hu and Linfeng Zhang},
  title         = {Token Pruning for Caching Better: 9 Times Acceleration on Stable Diffusion for Free},
  year          = {2024},
  url           = {https://arxiv.org/abs/2501.00375},
  archiveprefix = {arXiv},
  eprint        = {2501.00375},
  howpublished  = {arXiv preprint arXiv:2501.00375},
  primaryclass  = {cs.CV},
}

@inproceedings{sparsedit,
  author        = {Shuning Chang and Pichao Wang and Jiasheng Tang and Fan Wang and Yi Yang},
  title         = {SparseDiT: Token Sparsification for Efficient Diffusion Transformer},
  booktitle     = {Advances in Neural Information Processing Systems (NeurIPS)},
  year          = {2025},
  url           = {https://arxiv.org/abs/2412.06028},
  archiveprefix = {arXiv},
  eprint        = {2412.06028},
  primaryclass  = {cs.CV},
}

@inproceedings{tap,
    author    = {Zhu, Haowei and Huang, Tingxuan and Wang, Xing and Zhao, Tianyu and Wang, Jiexi and Chen, Weifeng and Peng, Xurui and Chen, Fangmin and Yong, Junhai and Wang, Bin},
    title     = {TAP: A Token-Adaptive Predictor Framework for Training-Free Diffusion Acceleration},
    booktitle = {Proceedings of the IEEE/CVF Conference on Computer Vision and Pattern Recognition (CVPR)},
    month     = {June},
    year      = {2026},
    pages     = {36082-36091}
}

@inproceedings{pab,
 author = {Zhao, Xuanlei and Jin, Xiaolong and Wang, Kai and You, Yang},
 booktitle = {International Conference on Learning Representations},
 editor = {Y. Yue and A. Garg and N. Peng and F. Sha and R. Yu},
 pages = {3296--3319},
 title = {Real-Time Video Generation with Pyramid Attention Broadcast},
 url = {https://proceedings.iclr.cc/paper_files/paper/2025/file/092c2d45005ea2db40fc24c470663416-Paper-Conference.pdf},
 volume = {2025},
 year = {2025}
}

@article{corgi,
  title={CorGi: Contribution-Guided Block-Wise Interval Caching for Training-Free Acceleration of Diffusion Transformers},
  author={Son, Yonglak and Kim, Suhyeok and Kim, Seungryong and Kim, Young Geun},
  journal={arXiv preprint arXiv:2512.24195},
  year={2025}
}

@inproceedings{soda,
  title={SODA: Sensitivity-Oriented Dynamic Acceleration for Diffusion Transformer},
  author={Shao, Tong and Fu, Yusen and Sun, Guoying and Kong, Jingde and Tian, Zhuotao and Su, Jingyong},
  booktitle={CVPR},
  year={2026}
}

@inproceedings{ditfastattn,
  author = {Yuan, Zhihang and Zhang, Hanling and Lu, Pu and Ning, Xuefei and Zhang, Linfeng and Zhao, Tianchen and Yan, Shengen and Dai, Guohao and Wang, Yu},
  booktitle = {Advances in Neural Information Processing Systems},
  doi = {10.52202/079017-0037},
  editor = {A. Globerson and L. Mackey and D. Belgrave and A. Fan and U. Paquet and J. Tomczak and C. Zhang},
  pages = {1196--1219},
  publisher = {Curran Associates, Inc.},
  title = {DiTFastAttn: Attention Compression for Diffusion Transformer Models},
  url = {https://proceedings.neurips.cc/paper_files/paper/2024/file/0267925e3c276e79189251585b4100bf-Paper-Conference.pdf},
  volume = {37},
  year = {2024}
}

@article{tgate,
  title={Cross-Attention Makes Inference Cumbersome in Text-to-Image Diffusion Models},
  author={Zhang, Wentian and Liu, Haozhe and Xie, Jinheng and Faccio, Francesco and Shou, Mike Zheng and Schmidhuber, J{\"u}rgen},
  journal={arXiv preprint arXiv:2404.02747v1},
  year={2024}
}

@inproceedings{clear,
  author        = {Liu, Songhua and Tan, Zhenxiong and Wang, Xinchao},
  title         = {CLEAR: Conv-Like Linearization Revs Pre-Trained Diffusion Transformers Up},
  booktitle     = {Advances in Neural Information Processing Systems (NeurIPS)},
  year          = {2025},
  url           = {https://arxiv.org/abs/2412.16112},
  archiveprefix = {arXiv},
  collection    = {NeurIPS 2025},
  eprint        = {2412.16112},
  primaryclass  = {cs.CV},
}

@misc{survey_lvlm_inference,
  author        = {Surendra Pathak and Bo Han},
  title         = {Towards Efficient Large Vision-Language Models: A Comprehensive Survey on Inference Strategies},
  year          = {2026},
  url           = {https://arxiv.org/abs/2603.27960},
  archiveprefix = {arXiv},
  eprint        = {2603.27960},
  howpublished  = {arXiv preprint arXiv:2603.27960},
  primaryclass  = {cs.LG},
}

@misc{ominicontrol2_arxiv25,
  author        = {Zhenxiong Tan and Qiaochu Xue and Xingyi Yang and Songhua Liu and Xinchao Wang},
  title         = {OminiControl2: Efficient Conditioning for Diffusion Transformers},
  year          = {2025},
  url           = {https://arxiv.org/abs/2503.08280},
  archiveprefix = {arXiv},
  eprint        = {2503.08280},
  howpublished  = {arXiv preprint arXiv:2503.08280},
  primaryclass  = {cs.CV},
}

@inproceedings{transPrune,
  title={TransPrune: Token Transition Pruning for Efficient Large Vision-Language Model},
  author={Li, Ao and Duan, Yuxiang and Zhang, Jinghui and Ma, Congbo and Xie, Yutong and Carneiro, Gustavo and Yaqub, Mohammad and Wang, Hu},
  booktitle={Proceedings of the IEEE/CVF Conference on Computer Vision and Pattern Recognition (CVPR)},
  year={2026}
}

@inproceedings{repreShift,
  title={Representation Shift: Unifying Token Compression with FlashAttention},
  author={Choi, Joonmyung and Lee, Sanghyeok and Ko, Byungoh and Kim, Eunseo and Kil, Jihyung and Kim, Hyunwoo J.},
  booktitle={Proceedings of the IEEE/CVF International Conference on Computer Vision},
  year={2025}
}

@inproceedings{att_flow_acl20,
  title={Quantifying attention flow in transformers},
  author={Abnar, Samira and Zuidema, Willem},
  booktitle={Proceedings of the 58th annual meeting of the association for computational linguistics},
  pages={4190--4197},
  year={2020}
}

@inproceedings{anycontrol,
  title={Anycontrol: create your artwork with versatile control on text-to-image generation},
  author={Sun, Yanan and Liu, Yanchen and Tang, Yinhao and Pei, Wenjie and Chen, Kai},
  booktitle={European Conference on Computer Vision},
  pages={92--109},
  year={2024},
  organization={Springer}
}

@article{pixelponder,
  title={Pixelponder: Dynamic patch adaptation for enhanced multi-conditional text-to-image generation},
  author={Pan, Yanjie and He, Qingdong and Jiang, Zhengkai and Xu, Pengcheng and Wang, Chaoyi and Peng, Jinlong and Wang, Haoxuan and Cao, Yun and Gan, Zhenye and Chi, Mingmin and others},
  journal={arXiv preprint arXiv:2503.06684},
  year={2025}
}

@inproceedings{unicombine,
  title={Unicombine: Unified multi-conditional combination with diffusion transformer},
  author={Wang, Haoxuan and Peng, Jinlong and He, Qingdong and Yang, Hao and Jin, Ying and Wu, Jiafu and Hu, Xiaobin and Pan, Yanjie and Gan, Zhenye and Chi, Mingmin and others},
  booktitle={Proceedings of the IEEE/CVF International Conference on Computer Vision},
  pages={18325--18334},
  year={2025}
}

@article{SCOPE,
  title={Scope: Saliency-coverage oriented token pruning for efficient multimodel llms},
  author={Deng, Jinhong and Li, Wen and Zhou, Joey Tianyi and He, Yang},
  journal={Advances in Neural Information Processing Systems},
  volume={38},
  pages={161527--161552},
  year={2026}
}

@article{glimpseprune,
  title={A glimpse to compress: Dynamic visual token pruning for large vision-language models},
  author={Zeng, Quan-Sheng and Li, Yunheng and Wang, Qilong and Jiang, Peng-Tao and Wu, Zuxuan and Cheng, Ming-Ming and Hou, Qibin},
  journal={IEEE Transactions on Circuits and Systems for Video Technology},
  year={2026},
  publisher={IEEE}
}

@inproceedings{LD_Pruner_cvpr24,
  title={Ld-pruner: Efficient pruning of latent diffusion models using task-agnostic insights},
  author={Castells, Thibault and Song, Hyoung-Kyu and Kim, Bo-Kyeong and Choi, Shinkook},
  booktitle={Proceedings of the IEEE/CVF Conference on Computer Vision and Pattern Recognition},
  pages={821--830},
  year={2024}
}

@article{Dip_go_nips23,
  title={Dip-go: A diffusion pruner via few-step gradient optimization},
  author={Zhu, Haowei and Tang, Dehua and Liu, Ji and Lu, Mingjie and Zheng, Jintu and Peng, Jinzhang and Li, Dong and Wang, Yu and Jiang, Fan and Tian, Lu and others},
  journal={Advances in Neural Information Processing Systems},
  volume={37},
  pages={92581--92604},
  year={2024}
}

@inproceedings{structural_pruning_nips23_maXinYin,
  title={Structural pruning for diffusion models},
  author={Gongfan Fang and Xinyin Ma and Xinchao Wang},
  booktitle={Advances in Neural Information Processing Systems},
  year={2023},
}

@inproceedings{MB,
    author    = {Oshima, Yuta and Miyake, Daiki and Matsutani, Kohsei and Iwasawa, Yusuke and Suzuki, Masahiro and Matsuo, Yutaka and Furuta, Hiroki},
    title     = {MultiBanana: A Challenging Benchmark for Multi-Reference Text-to-Image Generation},
    booktitle = {Proceedings of the IEEE/CVF Conference on Computer Vision and Pattern Recognition (CVPR)},
    month     = {June},
    year      = {2026},
    pages     = {448-460}
}

@article{macro,
  title     = {MACRO: Advancing Multi-Reference Image Generation with Structured Long-Context Data},
  author    = {Zhekai Chen and Yuqing Wang and Manyuan Zhang and Xihui Liu},
  journal   = {arXiv preprint arXiv:2603.25319},
  year      = {2026},
  url       = {https://arxiv.org/abs/2603.25319},
}

@article{qwen3VL,
      title={Qwen3-VL Technical Report}, 
      author={Shuai Bai and Yuxuan Cai and Ruizhe Chen and Keqin Chen and Xionghui Chen and Zesen Cheng and Lianghao Deng and Wei Ding and Chang Gao and Chunjiang Ge and Wenbin Ge and Zhifang Guo and Qidong Huang and Jie Huang and Fei Huang and Binyuan Hui and Shutong Jiang and Zhaohai Li and Mingsheng Li and Mei Li and Kaixin Li and Zicheng Lin and Junyang Lin and Xuejing Liu and Jiawei Liu and Chenglong Liu and Yang Liu and Dayiheng Liu and Shixuan Liu and Dunjie Lu and Ruilin Luo and Chenxu Lv and Rui Men and Lingchen Meng and Xuancheng Ren and Xingzhang Ren and Sibo Song and Yuchong Sun and Jun Tang and Jianhong Tu and Jianqiang Wan and Peng Wang and Pengfei Wang and Qiuyue Wang and Yuxuan Wang and Tianbao Xie and Yiheng Xu and Haiyang Xu and Jin Xu and Zhibo Yang and Mingkun Yang and Jianxin Yang and An Yang and Bowen Yu and Fei Zhang and Hang Zhang and Xi Zhang and Bo Zheng and Humen Zhong and Jingren Zhou and Fan Zhou and Jing Zhou and Yuanzhi Zhu and Ke Zhu},
	  journal={arXiv preprint arXiv:2511.21631},
      year={2025}
}

@article{birefnet,
  title={Bilateral Reference for High-Resolution Dichotomous Image Segmentation},
  author={Zheng, Peng and Gao, Dehong and Fan, Deng-Ping and Liu, Li and Laaksonen, Jorma and Ouyang, Wanli and Sebe, Nicu},
  journal={CAAI Artificial Intelligence Research},
  volume = {3},
  pages = {9150038},
  year={2024}
}

@article{RoPE,
  title={Roformer: Enhanced transformer with rotary position embedding},
  author={Su, Jianlin and Ahmed, Murtadha and Lu, Yu and Pan, Shengfeng and Bo, Wen and Liu, Yunfeng},
  journal={Neurocomputing},
  volume={568},
  pages={127063},
  year={2024},
  publisher={Elsevier}
}

@article{flashattention,
  title={Flashattention: Fast and memory-efficient exact attention with io-awareness},
  author={Dao, Tri and Fu, Dan and Ermon, Stefano and Rudra, Atri and R{\'e}, Christopher},
  journal={Advances in neural information processing systems},
  volume={35},
  pages={16344--16359},
  year={2022}
}

@misc{lmdeploy,
    title={LMDeploy: A Toolkit for Compressing, Deploying, and Serving LLM},
    author={LMDeploy Contributors},
    howpublished = {\url{https://github.com/InternLM/lmdeploy}},
    year={2023}
}

@misc{catPrune,
  author        = {Xinle Cheng and Zhuoming Chen and Zhihao Jia},
  title         = {CAT Pruning: Cluster-Aware Token Pruning For Text-to-Image Diffusion Models},
  year          = {2025},
  url           = {https://arxiv.org/abs/2502.00433},
  archiveprefix = {arXiv},
  eprint        = {2502.00433},
  howpublished  = {arXiv preprint arXiv:2502.00433},
  primaryclass  = {cs.CV},
}

@article{A2Edit,
  title={A$^2$-Edit: Precise Reference-Guided Image Editing of Arbitrary Objects and Ambiguous Masks},
  author={Zheng, Huayu and Li, Guangzhao and Zhao, Baixuan and Luo, Siqi and Jiang, Hantao and Zhai, Guangtao and Liu, Xiaohong},
  journal={arXiv preprint arXiv:2603.10685},
  year={2026}
}

@inproceedings{humof,
  title={HUMOF: Human motion forecasting in interactive social scenes},
  author={Caiyi Sun and Yujing Sun and Xiao Han and Zemin Yang and Jiawei Liu and Xinge Zhu and Siu Ming Yiu and Yuexin Ma},
  booktitle={International Conference on Learning Representations},
  volume={2026},
  pages={6511--6534},
  year={2026}
}

@inproceedings{c2f_c2fTrans_posenet,
  title={Efficient posenet with coarse to fine transformer},
  author={Li, Shaohua and Zhang, Haixiang and Ma, Hanjie and Feng, Jie and Jiang, Mingfeng},
  booktitle={ICASSP 2024-2024 IEEE International Conference on Acoustics, Speech and Signal Processing (ICASSP)},
  pages={5100--5104},
  year={2024},
  organization={IEEE}
}

@inproceedings{c2f_3d_highRes__magic3d,
  title={Magic3d: High-resolution text-to-3d content creation},
  author={Lin, Chen-Hsuan and Gao, Jun and Tang, Luming and Takikawa, Towaki and Zeng, Xiaohui and Huang, Xun and Kreis, Karsten and Fidler, Sanja and Liu, Ming-Yu and Lin, Tsung-Yi},
  booktitle={2023 IEEE/CVF Conference on Computer Vision and Pattern Recognition (CVPR)},
  pages={300--309},
  year={2023},
  organization={IEEE}
}

@article{c2f_video_highRes,
  title={FlashVSR: Towards Real-Time Diffusion-Based Streaming Video Super-Resolution},
  author={Zhuang, Junhao and Guo, Shi and Cai, Xin and Li, Xiaohui and Liu, Yihao and Yuan, Chun and Xue, Tianfan},
  journal={arXiv preprint arXiv:2510.12747},
  year={2025}
}

@article{draftAtt,
  title={DraftAttention: Fast Video Diffusion via Low-Resolution Attention Guidance},
  author={Shen, Xuan and Han, Chenxia and Zhou, Yufa and Xie, Yanyue and Gong, Yifan and Wang, Quanyi and Wang, Yiwei and Wang, Yanzhi and Zhao, Pu and Gu, Jiuxiang},
  journal={arXiv preprint arXiv:2505.14708},
  year={2025}
}

@article{lot,
  title={Level-of-Token Diffusion},
  author={Nakayama, Kiyohiro and Chao, Brian and Ackermann, Jan and Chen, Hansheng and Tombari, Federico and Guibas, Leonidas and Yariv, Lior and Wetzstein, Gordon},
  journal={arXiv preprint arXiv:2610.05816},
  year={2026}
}

@article{budgetpix,
  title={BudgetPix: Compute-Adaptive Tokenization for Pixel-Space Image Diffusion},
  author={Ozgur Kara and Yujia Chen and Daniel Watson and David Forsyth and James Matthew Rehg and Wen-Sheng Chu and Du Tran},
  journal={arXiv preprint arXiv:2610.12307},
  year={2026}
}

\appendix

The appendix is organized as follows:
\begin{itemize}
    \item Section~\ref{sec:suppl_analysis} presents additional analysis:
    \begin{itemize}
        \item Sec.~\ref{sec:theo_flops}: Theoretical FLOPs analysis.
        \item Sec.~\ref{sec:suppl_orth}: Orthogonality with existing acceleration techniques.
        \item Sec.~\ref{sec:suppl_nref}: Results under varying numbers of reference images.
        \item Sec.~\ref{sec:suppl_diff_more_RW}: Discussions with broader works.
    \end{itemize}

    \item Section~\ref{sec:suppl_impl_detail} provides implementation details:
    \begin{itemize}
        \item Sec.~\ref{sec:suppl_models}: Models and inference.
        \item Sec.~\ref{sec:suppl_decision}: Reduction decision policies.
        \item Sec.~\ref{sec:suppl_data}: Datasets and evaluation.
        \item Sec.~\ref{sec:suppl_compared_SOTA}: Details of compared methods.
        \item Sec.~\ref{sec:suppl_ear}: Effective Attention Ratio (EAR).
    \end{itemize}

    \item Section~\ref{sec:limitation} discusses the limitations of our work.

\end{itemize}

\section{Additional Analysis}
\label{sec:suppl_analysis}

\subsection{Theoretical Complexity Analysis}
\label{sec:theo_flops}

We analyze the theoretical FLOPs of the probing stage and the formal denoising stage, and derive the break-even reduction ratio at which the end-to-end cost matches that of the base model (i.e., theoretical speedup $=1$).
Since the attention and FFN modules dominate the computational complexity, we only analyze their contributions to the computational cost here.

\paragraph{Per-layer cost.}
Let $N$ be the joint sequence length and $d$ the embedding dimension.
We assume the FFN intermediate width is $4d$ (a standard dense-Transformer setting).
A common accounting then gives
\begin{equation}
C(N)
=
\underbrace{4\,N^{2}d}_{\text{attention}}
+
\underbrace{24\,N d^{2}}_{\text{QKVO + FFN}}.
\label{eq:layer_cost}
\end{equation}
Here we count FLOPs as floating-point operations, including both multiplications and additions: each matrix product of size $(N{\times}d){\times}(d{\times}N)$ costs $N^2d$ multiplications and as many additions, giving the attention term $2\,N^2d$ (for $QK^{\top}$) $+\,2\,N^2d$ (for $AV$) $=\,4\,N^2d$; the Q/K/V/O projections each cost $2Nd^2$ ($8Nd^2$ in total), and the FFN's two linear layers ($d\to4d$ and $4d\to d$) each cost $2Nd{\cdot}4d=8Nd^2$, i.e., $16Nd^2$ in total.
With $T$ denoising steps and $L$ layers, the baseline cost is $TL\,C(N)$.

\paragraph{Sequence lengths.}
Let $N_{P}$, $N_{G}$, and $N_{R}$ denote the numbers of prompt, generative, and reference tokens, so $N=N_{P}+N_{G}+N_{R}$.
In the probing stage, images are spatially downsampled by $2\times$ (vision tokens become $1/4$) and the temporal schedule is downsampled by a factor $D$. We instantiate $D \in \{2, 5\}$.
Thus
$$
N'
=
N_{P}
+
\tfrac{1}{4}N_{G}
+
\tfrac{1}{4}N_{R}.
$$
Let $\text{reduction\_{}ratio} \in [0, 1]$ denote the fraction of reference tokens removed after token reduction. The formal-stage length is
\begin{equation}
N_{f} = N_{P} + N_{G} + (1-\text{reduction\_{}ratio})\,N_{R}.
\end{equation}
The total cost of our method is therefore
\begin{equation}
\frac{T}{D}\,L\,C(N') + T\,L\,C(N_{f}).
\end{equation}

\paragraph{Break-even condition.}
Matching the baseline requires
\begin{equation}
\frac{1}{D}\,C(N')+C(N_{f})=C(N).
\label{eq:breakeven}
\end{equation}

\paragraph{Numerical instantiation.}
We instantiate a representative two-reference setting with $N_{G}=N_{R_{1}}=N_{R_{2}}=4096$ (so $N_{R}=8192$), $N_{P}=100$, and $d=3072$.
Then $N=12388$, $N'=3172$, and solving~\eqref{eq:breakeven} yields a break-even ratio of approximately $9.9\%$ for $D=2$ (for 4-step denoising model like Flux.2-Klein) and $3.9\%$ for $D=5$ (e.g., 50-step model HiDream-O1-Image).
This break-even threshold is quite low that it is almost always satisfied in practice, meaning the savings from token reduction easily outweigh the probing overhead, yielding net end-to-end speedup.

\subsection{Orthogonality with Most Existing Acceleration Techniques}
\label{sec:suppl_orth}
Our method is naturally orthogonal and complementary to existing acceleration methods from two perspectives:
\begin{itemize}
    \item \textbf{Stage-level Compatibility:} Most existing acceleration methods operate as \textit{in-model token reduction}, which prunes or merges tokens inside the model during the main denoising loop. In contrast, our method operates outside the main denoising loop and reduces reference tokens before the main denoising stage begins. It directly provides a standard, shortened token sequence to the base model without altering internal model structures or operations, allowing existing in-model methods to be directly applied on top of our reduced sequence.
    \item \textbf{Redundancy-level Orthogonality:} Our method specifically addresses the \textit{spatial token redundancy} in reference images. This is naturally orthogonal to methods focusing on \textit{temporal redundancy} (such as cross-step feature caching and distillation) or methods that prune or sparsify generative tokens (like V-V sparse attention in video generation). 
\end{itemize}
Consequently, our early token reduction can be seamlessly combined with these existing acceleration techniques to achieve compounded efficiency improvement in diffusion-based visual generation tasks.

\paragraph{Clarification on Terminology:}
In literature, methods compressing tokens before the main network backbone are broadly termed \textbf{``pre-model''} token reduction. In this paper, we primarily refer to our method as \textbf{``early''} token reduction to avoid ambiguity. 
Since our framework executes a lightweight downsampled self-probing stage using the DiT model itself to extract importance scores, the DiT model has already been executed before the token reduction takes place. Therefore, strictly calling it ``pre-model'' might be misleading. We instead adopt the term ``early'' token reduction to precisely denote that reference tokens are reduced at an early stage (prior to the main denoising stage).

In addition to the default timestep-mapped reduction (Sec~\ref{sec:method_reduction}), our framework seamlessly supports reducing reference tokens just once for the entire main denoising loop. By averaging the reference utilization scores across all probing steps, tokens can be filtered once before the main loop begins. 
Note that both modes share the same fundamental principle: reference tokens are always pruned or substituted outside the DiT in the main denoising, requiring zero structural or operational changes to the DiT in the main loop. 
We default (all experiments in the paper) to timestep-mapped reduction simply because matching the reference retention to the evolving demands across different denoising stages offers a slight performance gain.

\subsection{Results under varying numbers of reference images}
\label{sec:suppl_nref}

\begin{figure*}[h]
  \begin{center}
    \includegraphics[width=  0.7  \textwidth]{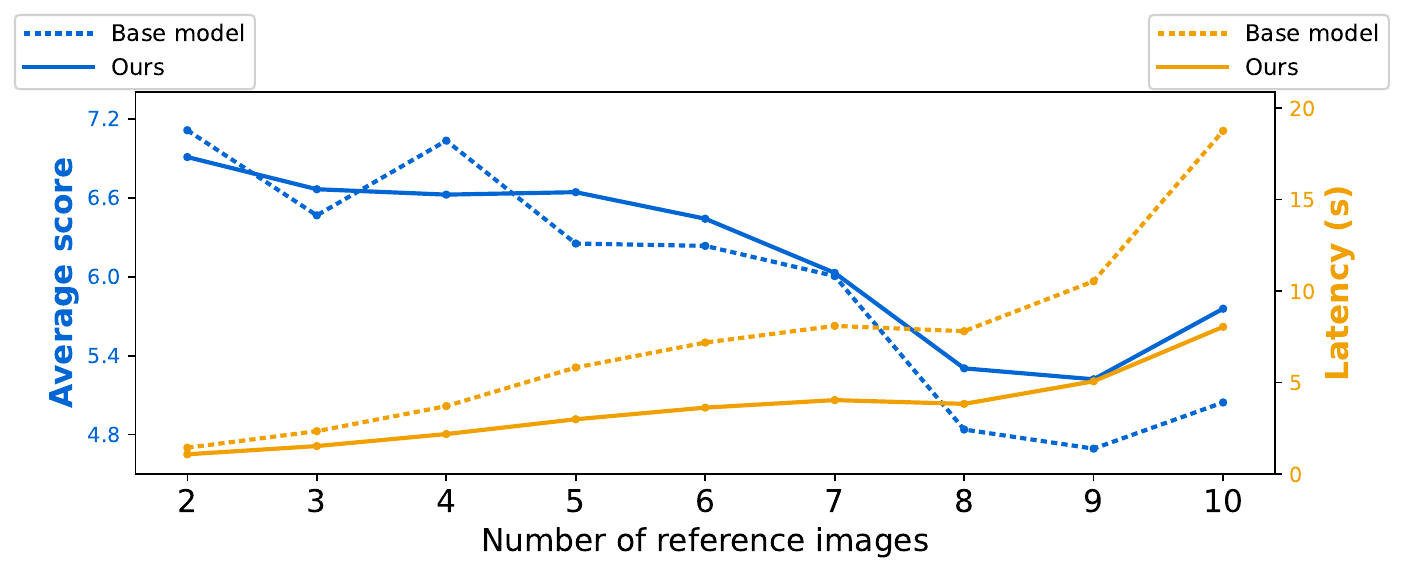}
  \end{center}
  \caption{
Average score and latency of Ours (slow) (Flux.2-Klein) on MACRO with 2--10 reference images.
}\label{fig:nref_score_latency}
\end{figure*}

Figure~\ref{fig:nref_score_latency} reports the score and latency of Ours (slow) on MACRO with 2--10 reference images.

\subsection{Discussions with Broader Works}
\label{sec:suppl_diff_more_RW}

\paragraph{Relation to Dynamic Resolution and Adaptive Tokenization.}
The most recent methods LoT Diffusion~\citep{lot} and BudgetPix~\citep{budgetpix} introduce adaptive tokenization by assigning variable-sized patches based on visual complexity. 
Fundamentally, our tri-tier reduction strategy (preserving full-resolution tokens for critical details and substituting mid-importance regions with downsampled tokens) \textbf{also functions as a dynamic resolution-based efficient encoding}. 
Our token importance and multi-resolution layout are directly determined by the DiT itself through cheap downsampled self-probing in a fully training-free manner.

\paragraph{The Coarse-to-Fine Paradigms.}
Our downsampled self-probing mechanism aligns with the coarse-to-fine philosophy widely adopted across diverse fields outside diffusion acceleration, such as coarse-to-fine Transformer~\citep{c2f_c2fTrans_posenet,humof}, and high-resolution generation~\citep{c2f_3d_highRes__magic3d,c2f_video_highRes};
Sparse and $\mathcal{O}(N \log N)$ attention mechanisms~\citep{deepseek_NSA, zhou2026trainable_nlog,draftAtt} also adopt a comparable rationale, utilizing coarse-grained block selection or hierarchical structures to guide fine token-level interactions. Crucially, while they operate at the \textbf{attention level}, our framework instantiates a \textbf{model-level} coarse-to-fine paradigm: we first deploy a coarse run to probe reference token sparsity upfront, which then governs the token allocation for the subsequent fine run.

\section{Implementation Details}
\label{sec:suppl_impl_detail}

Our \textbf{code} will be made publicly available upon acceptance, containing all implementation details.

\subsection{Models and Inference}
\label{sec:suppl_models}

We evaluate on three base models:
the latent-space Diffusion Transformers (DiTs) Flux.2-Klein-4B and Qwen-Image-Edit-2511,
and
HiDream-O1-Image, a Pixel-level Unified Transformer (UiT) that directly processes raw pixels without a VAE.
Evaluating on these two architectures confirms our method is effective for diverse generic reference-based generation models.
For all base models, we use their official code and default configurations.
Specifically, we use $4$, $40$, and $50$ inference steps for Flux.2-Klein-4B~\cite{flux2}, Qwen-Image-Edit-2511~\cite{qwenImage}, and HiDream-O1-Image~\cite{hiDream}, respectively.
For the probing stage, let $\tilde{T} = T / D$ denote the number of probing denoising steps.
We set $\tilde{T} = 2$ ($D = 2$) for Flux.2-Klein-4B, $\tilde{T} = 10$ ($D = 4$) for Qwen-Image-Edit, and $\tilde{T} = 10$ ($D = 5$) for HiDream-O1-Image.
$\texttt{true\_cfg\_scale}=4.0$ for Qwen-Image-Edit-2511;
$\texttt{guidance\_scale}=1.0$ for Flux.2-Klein-4B;
$\texttt{guidance\_scale}=5.0$ for HiDream-O1-Image.
FlashAttention~\cite{flashattention} is used as the attention backend.
Since the attention matrix is not materialized under FlashAttention, we use a specialized Triton kernel to extract the reference utilization score.
VAE slicing and tiling are enabled when inferring on samples that encounter out of memory during the VAE encoding stage.
For each model, we report a quality-oriented \textbf{Ours(slow)} and a speed-oriented \textbf{Ours(fast)} by adjusting $\tau_1$ and $\tau_2$.
All experiments run on a single NVIDIA H800 (80GB) GPU.

For MultiBanana~\cite{MB}, which does not prescribe the input reference or output target size, we use each reference at its native resolution, except that any image whose area exceeds $\mathrm{MAX\_AREA}=4096\times4096$ is resized to this area budget; the last reference's size is used as the target generation size.
For Macro~\cite{macro}, we follow its official settings.
All runs use seed $42$.

\paragraph{Handling of Rotary Positional Embeddings (RoPE)}
Since the image generation models~\cite{flux2,qwenImage,hiDream} encode spatial layouts via RoPE~\cite{RoPE}, we maintain geometric consistency for the pruned reference sequence.
The preserved full-resolution tokens retain their original 2D RoPE coordinates.
For the replaced low-resolution tokens representing mid-importance blocks, we assign their RoPE coordinates precisely to the spatial center of the corresponding $2 \times 2$ block.

\subsection{Reduction Decision Policies}
\label{sec:suppl_decision}

\subsubsection{Ratio-based Decision (Default).}
In our experiments, we adopt a ratio-based policy.
Under this setting, retention proportions $p_1$ and $p_2$ are explicitly designated for the Full-Res Tier and Low-Res Tier, respectively. The cutoffs $\tau_1$ and $\tau_2$ in Eq.~(1) are dynamically determined on a per-sample basis: after sorting all downsampled reference token scores $\{\tilde{I}_{\tilde{r}}^{\tilde{t}}\}$ at probe timestep $\tilde{t}$ of the current sample in descending order, $\tau_1$ is set to the score at the top-$p_1$ percentile rank, and $\tau_2$ is set to the score at the top-$(p_1 + p_2)$ percentile rank. Consequently, the top-$p_1$ fraction enters the Full-Res Tier, the subsequent $p_2$ fraction is substituted into the Low-Res Tier, and the remaining tokens are discarded. The primary advantage of this policy is budget predictability, providing deterministic sequence lengths.

In Table~\ref{tab:mainB}, we configure two operational modes for this setting: \textbf{Ours(fast)} and \textbf{Ours(slow)}.
For \textbf{Ours(fast)}, we set the Full-Res Tier ratio to $p_1 = 15\%$ and the Low-Res Tier ratio to $p_2 = 2 \times p_1 = 30\%$.
For \textbf{Ours(slow)}, both proportions are doubled, setting $p_1 = 30\%$ and $p_2 = 60\%$.
These proportions represent global budgets evaluated across all reference images for a given sample. To prevent severe inter-reference imbalance—such as a visually dominant reference image exhausting all available quotas and starving other references—we enforce a minimum retention guarantee on each individual reference image. Specifically, every reference image is constrained to retain at least $\frac{2}{3}$ of the global Full-Res ratio within its own token grid, yielding a per-reference lower bound of $p_{\text{min}} = \frac{2}{3} p_1$ (i.e., at least $10\%$ for Ours(fast) and $20\%$ for Ours(slow)).

\paragraph{Timestep-Aware Ratio Shift.}
Because DiTs prioritize global layout planning in early steps and fine-grained detail refinement in late steps, high-resolution details are less critical early on, whereas a wider spatial context is more beneficial. We therefore dynamically modulate the retention proportions across denoising timesteps: during the earliest steps, the effective Full-Res ratio is scaled to $0.8 p_1$, while the Low-Res ratio is expanded to $1.2 p_2$. As generation proceeds toward the final refinement steps, these scaling factors linearly transition back to $1.0$ (restoring the base ratios $p_1$ and $p_2$). This dynamic adjustment slightly improves generation quality, so we adopt it in our default configuration.

\subsubsection{Threshold-based Decision.}
We also support a threshold-based decision policy for sample-difficulty awareness, where fixed absolute thresholds $\tau_1$ and $\tau_2$ are applied directly to the normalized reference utilization score $\tilde{I}_{\text{norm}}$. Before thresholding, we normalize the raw utilization score by the total number of downsampled reference tokens of the current sample, $\tilde{I}_{\text{norm}} = \tilde{I} \cdot N_{\tilde{R}}$, making the score agnostic to varying reference counts and resolutions across samples. 
The primary benefit of this strategy is sample-difficulty awareness: for complex reference scenes requiring rich context, the model naturally allocates high attention over broader regions, automatically preserving a larger token budget; simpler references with clean backgrounds are pruned more aggressively.

\subsection{Datasets and Evaluation}
\label{sec:suppl_data}

\textbf{MultiBanana}~\cite{MB} is a challenging benchmark for multi-reference image generation,
consisting of 3K+ samples with up to 8 reference images.
For the two-reference setting, it covers 11 editing tasks including subject addition, replacement, background change, color/material/pose/hairstyle/makeup modification, tone transformation, style transfer, and text correction.
For the $\ge 3$ references setting, 
it includes four tasks: object composition, background composition, local attribute transfer, and global tone edit.
The benchmark incorporates 
four difficulty axes: cross-domain diversity (e.g., mixing photos, anime, and sketches), scale and viewpoint differences, rare concept references (e.g., a red banana), and multilingual references (English, Chinese, and Japanese), comprehensively evaluating the model's ability on multi-reference generation tasks.
MultiBanana uses a VLM to score generation quality from five dimensions: Text-Instruction Alignment, Reference Consistency, Background-Subject Match, Physical Realism, and Visual Quality, each on a 10-point scale.
To align with our focus on multi-reference efficiency setting, our main experiment targets all multi-reference tasks (excluding the pure instruction-based editing task in MultiBanana).
We report the mean across tasks for each quality dimension.

To reduce decoding randomness, each sample is scored twice under different decoding configurations and the two scores are averaged:
(1) greedy decoding ($\texttt{temperature}=0$ and $\texttt{top\_k}=1$);
(2) Top-p sampling with $\texttt{temperature}=0.8$, $\texttt{top\_p}=0.8$, and $\texttt{random\_seed}=0$.

Following the recommendation in Issue~\#2 of the official MultiBanana GitHub repository, we exclude the local attribute
transfer samples with IDs $\le 200$,
since the maintainers confirmed a prompt-reference mismatch bug on these samples and advised excluding them from evaluation.

\textbf{Macro}~\cite{macro} is also a reference-based image generation dataset 
covering four tasks:
customization (composing multiple subjects into coherent scenes), illustration (generating narrative images based on interleaved multimodal context), spatial synthesis (novel view synthesis), and temporal synthesis (predicting a future keyframe from history keyframes), 
each comprising 1000 samples with 1--10 reference images, in total 4000 samples.
Macro evaluates each sample across multiple dimensions to compute an overall score. 
We report the average overall score across all samples.

Qwen3-VL~\cite{qwen3VL}, deployed via the LMDeploy~\cite{lmdeploy} framework on an NVIDIA H800 (80GB), is used as the VLM-as-judge for the two benchmarks.

For efficiency, we report end-to-end latency (including the VAE and probing denoising stage, excluding disk I/O such as reference image reading and output image dumping), peak activation VRAM across samples, and speedup over the corresponding base model.

\subsection{Details of Compared Methods}
\label{sec:suppl_compared_SOTA}
\paragraph{ToMeSD.}
For ToMeSD, we directly apply the official method to the reference image tokens before entering the DiT.
\paragraph{VisPruner.}
VisPruner~\cite{visPruner} first extracts an importance score for each visual token from a vision encoder’s self-attention. If the encoder has a CLS token, importance is the CLS-to-patch attention; if not (e.g., SigLIP), it averages attention over the query axis. It then keeps a budget of tokens by selecting the top-scoring important subset and filling the rest via iterative cosine-similarity deduplication for diversity.
For HiDream-O1-Image, we reuse the model’s built-in Qwen3-VL vision encoder to score reference tokens.
For FLUX.2 Klein, the model has no vision encoder, so we follow VisPruner’s default and use CLIP (ViT-L/14-336) for importance.
Because CLIP patches do not match Flux’s VAE-packed reference tokens, we compute CLS-to-patch attention at CLIP’s native resolution and bilinearly map the score map onto the Flux reference grid, yielding a spatially aligned importance map for subsequent stages. Important and diverse token selection then proceeds as in VisPruner, with cosine deduplication on the packed VAE latents that enter the DiT.
For Qwen-Image-Edit, we reuse the built-in Qwen2.5-VL vision encoder to score tokens. Importance is computed on the pipeline’s $384 \times 384$ condition views and bilinearly mapped onto the VAE-packed reference grid; VisPruner then selects important and diverse tokens with cosine deduplication on those packed VAE latents.

\subsection{Effective Attention Ratio (EAR)}
\label{sec:suppl_ear}

The Effective Attention Ratio (EAR) in Sec.~\ref{sec:EAR} is only possible when the ground-truth important region is known. While general reference-based tasks do not mark it, the object composition task in MultiBanana~\cite{MB} does.
Since the object composition task aims to composite the salient object in each reference image, the ground-truth important region is the salient object region in each reference image.
We obtain this salient region with BiRefNet~\cite{birefnet}, which returns a soft saliency map $s\in[0,1]$. The map is resized to the latent grid of the reference, and $(w,h)$ indexes a reference token on that grid.

With the region fixed, we measure the attention that generative tokens $G$ and prompt tokens $P$ place on each reference token inside the main denoising loop. At timestep $t$ and layer $l$, their total attention to the token at $(w,h)$ is $\sum_{i \in G \cup P} A_{i,(w,h)}^{t,l}$, where $A_{i,(w,h)}^{t,l}$ is the head-averaged attention from query $i$. Summing over timesteps and layers, the Effective Attention Ratio (EAR) is the share of this attention that lands on the salient region,
\begin{equation}
\text{Effective Attention Ratio (EAR)}
=
\frac{
\sum_{w,h}
\Big(
\sum_{t}\sum_{l}\sum_{i \in G \cup P} A_{i,(w,h)}^{t,l}
\Big)\, s_{(w,h)}
}{
\sum_{w,h}
\sum_{t}\sum_{l}\sum_{i \in G \cup P} A_{i,(w,h)}^{t,l}
}.
\end{equation}

Table~\ref{tab:ear} compares the unpruned base model and three ways of building the reference utilization score that drives token reduction. 
The quality scores in Table~\ref{tab:ear} are computed over the same samples as in Table~\ref{tab:mainB} and Figure~\ref{fig:comp}, whereas EAR is calculated on the object composition subset, as importance GT is only defined for this task.
The reduction budget is the same in all four: the same ratio-based rule fixes how many tokens are preserved and how many are substituted.
(1)~\textit{Baseline (No Reduction)} keeps the full reference sequence, so the main denoising sees every reference token.
(2)~\textit{Top-1 layer} reduces tokens with the score of the single layer best aligned with the oracle importance $I_r^*$. On Flux.2-Klein-4B~\cite{flux2} this is layer $l{=}15$, shown in Fig.~\ref{fig:attn_vis}(b), with $I_r^l = \frac{1}{T}\sum_t \sum_{i \in G \cup P} A_{i, r}^{t, l}$.
(3)~\textit{Top 20\% of layers} uses the mean of the top 20\% best-correlated middle layers. On Flux.2-Klein-4B these are $\mathcal{L}=\{14,15,16,17,18\}$, and $I_r = \frac{1}{|\mathcal{L}|}\sum_{l \in \mathcal{L}} I_r^l$.
(4)~\textit{All layers} uses the mean over every transformer block, $I_r = \frac{1}{L}\sum_{l=1}^{L} I_r^l = \frac{1}{LT}\sum_{l=1}^{L}\sum_t \sum_{i \in G \cup P} A_{i, r}^{t, l}$.

\section{Limitations}
\label{sec:limitation}

1.
This paper focuses on the efficiency drop caused by token growth when multiple reference images are used.
In the single-image setting, namely pure instruction-based editing, the acceleration gain is relatively lower than in the multi-reference case.

2.
The empirical validation of our framework is currently limited to multi-reference-based image generation. In principle, the underlying strategy of early downsampled self-probing to identify spatial redundancy can be transferred to \textbf{broader visual tasks} like ViTs, VLMs, and long video generation.

\end{document}